%% file: main.tex
\documentclass{article}

\usepackage{microtype}
\usepackage{graphicx}
\usepackage{subcaption}
\usepackage{booktabs} 
\usepackage{multirow}
\usepackage{makecell}

\usepackage{hyperref}

\usepackage[preprint]{icml2026}

\usepackage{amsmath}
\usepackage{amssymb}
\usepackage{mathtools}
\usepackage{amsthm}
\usepackage{tabularx}

\usepackage[capitalize,noabbrev]{cleveref}

\theoremstyle{plain}

\theoremstyle{definition}

\theoremstyle{remark}

\usepackage[textsize=tiny]{todonotes}

\newcommand{\methodname}{\textbf{FBCIR}}
\newcommand{\dataname}{\textbf{FBCIR-Data}}
\definecolor{green}{RGB}{0,180,30}
\definecolor{red}{RGB}{200,0,0}

\icmltitlerunning{\methodname: Balancing Cross-Modal Focuses in Composed Image Retrieval}

\begin{document}

\twocolumn[
  \icmltitle{\methodname: Balancing Cross-Modal Focuses in Composed Image Retrieval}
  
  \icmlsetsymbol{equal}{*}

  \begin{icmlauthorlist}
    \icmlauthor{Chenchen Zhao}{cuhk,tencent}
    \icmlauthor{Jianhuan Zhuo}{tencent}
    \icmlauthor{Muxi Chen}{cuhk,tencent}
    \icmlauthor{Zhaohua Zhang}{tencent,dlut}
    \icmlauthor{Wenyu Jiang}{tencent,nju}\\
    \icmlauthor{Tianwen Jiang}{tencent}
    \icmlauthor{Qiuyong Xiao}{tencent}
    \icmlauthor{Jihong Zhang}{tencent}
    \icmlauthor{Qiang Xu}{cuhk}
  \end{icmlauthorlist}

  \icmlaffiliation{cuhk}{Department of Computer Science and Engineering, The Chinese University of Hong Kong}
  \icmlaffiliation{dlut}{School of Mathematical Sciences, Dalian University of Technology}
  \icmlaffiliation{nju}{School of Computer Science, Nanjing University}
  \icmlaffiliation{tencent}{Tencent AI Data Department}

  \icmlcorrespondingauthor{Qiang Xu}{qxu@cse.cuhk.edu.hk}




  \icmlkeywords{Composed Image Retrieval, Benchmark, Model Interpretation}

  \vskip 0.3in
]



\printAffiliationsAndNotice{}  

\input{sections/abstract}
\input{sections/01intro}
\input{sections/02relatedwork}
\input{sections/03method}
\input{sections/04benchmark}
\input{sections/05experiments}
\input{sections/06limitations}
\input{sections/07conclusion}
\input{sections/impact}

\bibliography{main}
\bibliographystyle{icml2026}

\newpage
\appendix
\onecolumn

\input{sections/appendix}

\end{document}

%% file: sections/abstract.tex
\begin{abstract}
Composed image retrieval (CIR) requires multi-modal models to jointly reason over visual content and semantic modifications presented in text-image input pairs. While current CIR models achieve strong performance on common benchmark cases, their accuracies often degrades in more challenging scenarios where negative candidates are semantically aligned with the query image or text. In this paper, we attribute this degradation to \textit{focus imbalances}, where models disproportionately attend to one modality while neglecting the other. To validate this claim, we propose \methodname, a multi-modal focus interpretation method that identifies the most crucial visual and textual input components to a model's retrieval decisions. Using \methodname, we report that focus imbalances are prevalent in existing CIR models, especially under hard negative settings. Building on the analyses, we further propose a CIR data augmentation workflow that facilitates existing CIR datasets with curated hard negatives designed to encourage balanced cross-modal reasoning. Extensive experiments across multiple CIR models demonstrate that the proposed augmentation consistently improves performance in challenging cases, while maintaining their capabilities on standard benchmarks. Together, our interpretation method and data augmentation workflow provide a new perspective on CIR model diagnosis and robustness improvements.
\end{abstract}

%% file: sections/01intro.tex
\section{Introduction}\label{sec:intro}

Image retrieval (IR) has long been a fundamental problem in computer vision~\cite{isola2015discovering} and vision-language research~\cite{liu2021image}. Traditional IR methods with uni-modal paradigms (i.e., queries are specified either by images or by text) are limited in addressing increasingly complex user requirements that involve multi-modal specifications. Composed image retrieval (CIR)~\cite{vo2019composing} addresses this limitation by enabling multi-modal queries (images \& texts in most cases). Given an image and a related modifying instruction, CIR aims to retrieve images that are semantically consistent with both the visual and the textual content. Early CIR methods~\cite{baldrati2022effective, saito2023pic2word, li2022blip} extend CLIP-like models~\cite{radford2021learning} with multi-modal input capabilities, while more recent methods~\cite{bai2023qwen} achieve improved performance by adopting vision-language models (VLMs).

\begin{figure}[t]
\centering
\includegraphics[width=\linewidth]{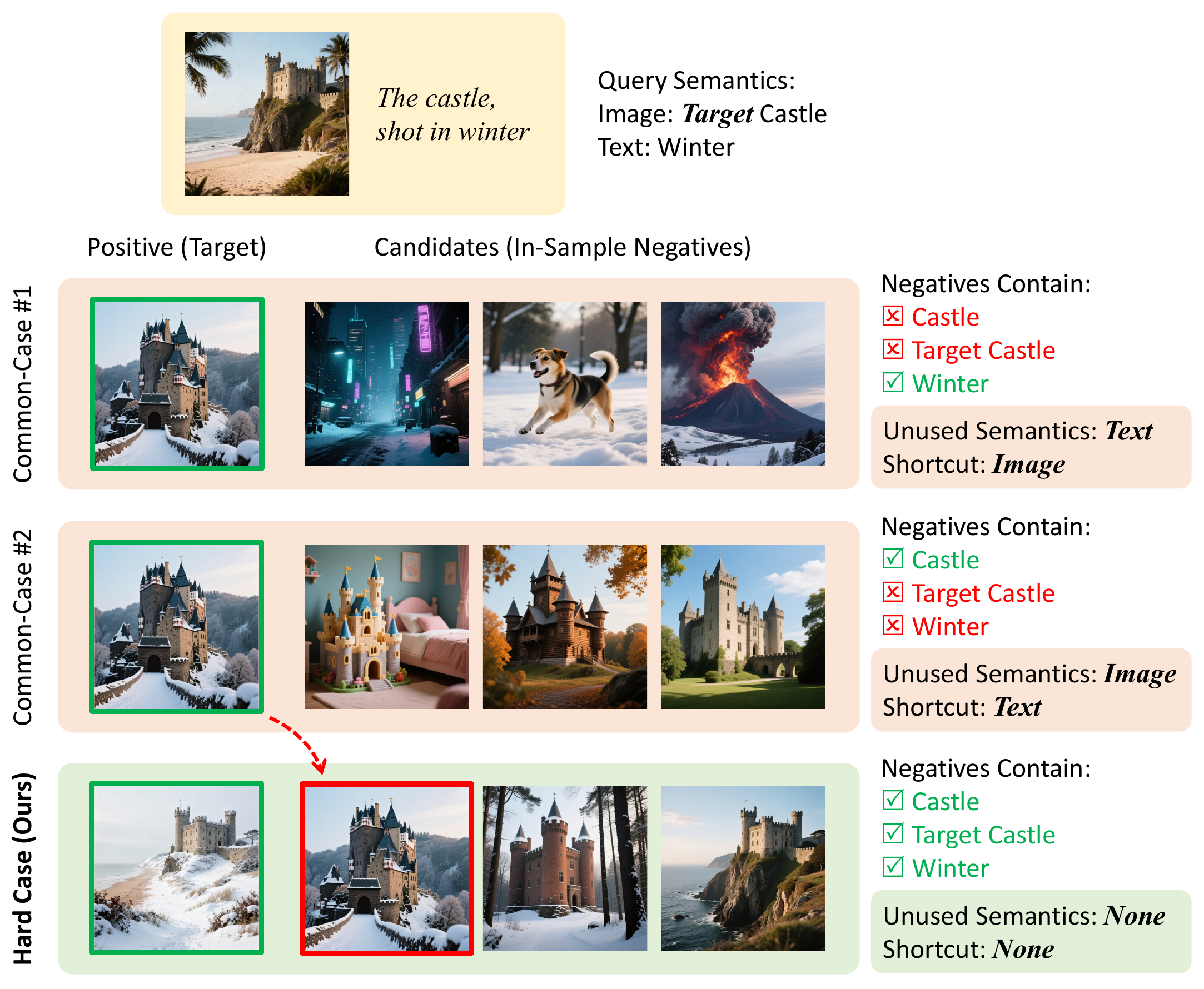}
\caption{Models trained on common-case data learn ``shortcuts'' to get correct common-case retrieval results, but tend to fail on hard cases that require balanced cross-modal focuses. In this paper, we try to solve this issue by constructing targeted hard negatives. Original positives that are not highly consistent with the query are also considered negative in the proposed framework and replaced by more consistent synthetic images.}
\vspace{-0.2cm}
\label{fig:problem}
\end{figure}

A central challenge in CIR is the requirement of jointly reasoning over semantics from both modalities. In many common cases, there are large semantic gaps between targets and negative candidates. This setting allows models to rely on ``shortcut'' strategies, producing correct retrieval results without equally processing information from both modalities, but generally fail in more challenging scenarios where negatives closely resemble targets and are semantically consistent with one of the input modalities. An intuitive example is shown in Figure~\ref{fig:problem}. In common-case \#1, all negative candidates lack the key visual concept (castle) required by the query image, enabling the model to correctly retrieve the target without referring to the text semantics. Similarly, all negatives in common-case \#2 lack the key textual concept (winter) specified by the text, enabling correct retrieval without referring to the image semantics. However, in harder cases where both modalities contribute essential but complementary information, such shortcut behavior often leads to retrieval failures. This difficulty is a key factor that makes CIR more challenging than conventional IR, and contributes to the generally lower performance of CIR models~\cite{jiang2024vlm2vec}. Additional analyses of such behavior and its impacts on model performance are detailed in Appendix~\ref{app:focus_analysis}.

\begin{figure*}[t]
\centering
\includegraphics[width=0.9\textwidth]{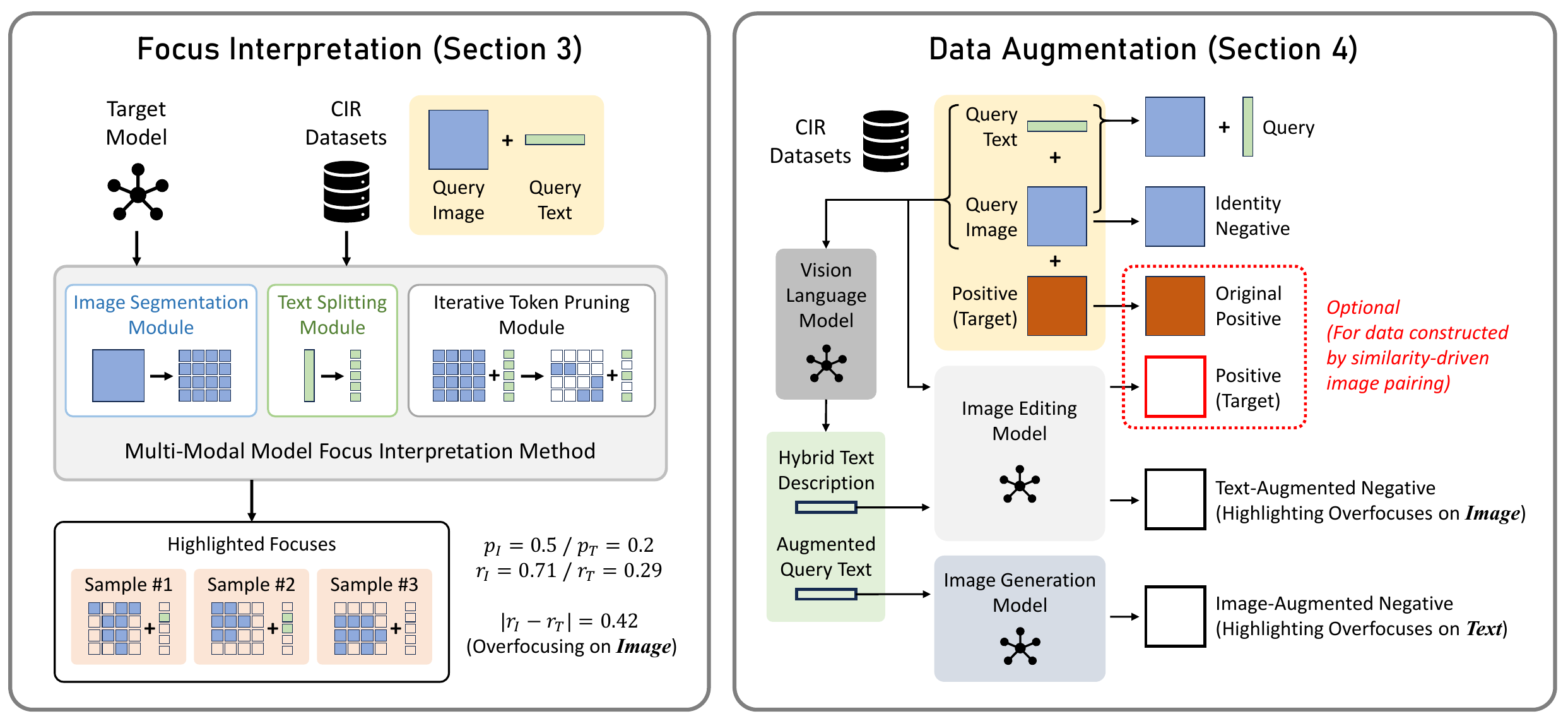}
\caption{The overall framework of \methodname, including a multi-modal model focus interpretation method and a dataset augmentation workflow. Given CIR triplets, the focus interpretation method highlights specific image segments and instruction keywords as the model's focuses, and reveals possible focus imbalances. The dataset augmentation framework facilitates existing CIR triplets with crafted hard negatives that encourage more balanced focuses. The focus interpretation module serves as the problem indicator and post-hoc validator of the data augmentation module.}
\vspace{-0.2cm}
\label{fig:teaser}
\end{figure*}

In this paper, we term such shortcuts \textit{focus imbalances}, where a model over-attends to one modality while neglecting the other.
To validate the prevalence and severity of this issue, we first introduce a multi-modal focus interpretation method, termed \methodname~(\textbf{\underline{F}}ocus-\textbf{\underline{B}}alancing). Using \methodname, we reveal that focus imbalances are common across existing CIR models. Building on the analyses, we further propose a CIR data augmentation workflow designed to mitigate focus imbalances. The \dataname~workflow systematically enriches existing CIR triplets with curated hard negatives that require balanced reasoning over both image and text semantics. Using the workflow, we construct a novel benchmark and a finetuning dataset that enable quantitative evaluations and targeted improvements of CIR models from the perspective of focus balancing. Extensive experiments demonstrate that the proposed workflow effectively improves model performance on challenging cases while alleviating focus imbalances. An overview of the \methodname~framework is shown in Figure~\ref{fig:teaser}.

The contributions of this work are three-fold:
\begin{itemize}
\item We identify and formalize the problem of \textit{focus imbalance} in CIR, where models exploit shortcut cues from a single modality instead of jointly reasoning over image and text semantics, and empirically demonstrate its prevalence across representative CIR models.
\item We propose \methodname, a multi-modal focus interpretation method that provides fine-grained analyses of multi-modal focus behavior, enabling systematic diagnosis and quantitative assessment of focus imbalances in CIR models.
\item We develop a data augmentation workflow based on \methodname, yielding a targeted hard-case benchmark and a finetuning dataset with curated hard negatives, which effectively improve models' hard-case performance and encourage more balanced cross-modal focuses.
\end{itemize}


%% file: sections/02relatedwork.tex
\section{Related Works}

\subsection{Composed Image Retrieval (CIR) Models}

Composed Image Retrieval (CIR) retrieves images by conditioning jointly on a reference image and a modifying textual description. Early methods~\cite{vo2019composing} focus on explicit feature composition over uni-modal or weakly aligned multimodal representations. With the development of vision-language pretraining, recent CLIP-based~\cite{radford2021learning,baldrati2022effective,saito2023pic2word} and VLM-based~\cite{bai2023qwen,jian2025rzenembed,lin2024mm} approaches employ more expressive fusion modules or dedicated query encoders, achieving strong performance on standard CIR benchmarks.

However, most CIR models are trained and evaluated under relatively easy task settings, where negative candidates are semantically distant from the target. This allows models to over-rely on a single modality rather than integrating multi-modal information. Such ``shortcuts'' have received limited explicit analysis, motivating in-depth investigation of model focuses under more challenging retrieval scenarios.

\subsection{CIR Datasets and Benchmarks}

Several benchmark datasets drive CIR research. FashionIQ~\cite{wu2021fashion} focuses on fashion images with modification instructions, while CIRR~\cite{liu2021image} extends to diverse real-world objects with relative captions. Earlier datasets~\cite{isola2015discovering} also support attribute-based or relative retrieval, but within more constrained domains.

Despite their impacts, these benchmarks generally rely on simple negative sampling strategies, making retrieval achievable even when models attend primarily to one modality. Although specific works~\cite{baldrati2022effective} introduce harder negatives or refined protocols, systematic evaluation of cross-modal focuses remains limited. In particular, existing datasets rarely include candidates that correlate strongly with one specific modality of the composed query, which are crucial for revealing focus imbalances. This work explicitly targets such scenarios and provides a complementary evaluation perspective beyond conventional retrieval accuracy.


\subsection{Model Focus Identification and Interpretation}

A line of research on model interpretation aims to identify input components most influential to model decisions. Gradient-based saliency methods~\cite{simonyan2013deep, selvaraju2017grad} and perturbation-based methods~\cite{zeiler2014visualizing,fong2017interpretable,ribeiro2016should,lundberg2017unified} analyze model behavior by attributing predictions to specific inputs or by observing output changes under controlled input modifications. Some works~\cite{focalogic,chang2018explaining} further identify minimal sufficient input regions that preserve model predictions, yielding deterministic and interpretable explanations.

While effective for single-modality vision tasks, existing interpretation are not designed to identify multi-modal model focuses, nor to account for cross-modal focus balances. This limitation motivates focus-aware analysis frameworks that jointly examine visual and textual contributions in CIR.

%% file: sections/03method.tex
\section{\methodname: A Multi-Modal Focus Interpretation Method}\label{sec:interpretation}

\methodname~is a model behavior interpretation method from the ``focus'' perspective. Given a model and a validation dataset, it identifies focuses of the model on the query images and texts, enabling quantitative measurement of cross-modal focus imbalances. This allows direct validation of the issue stated in Section~\ref{sec:intro} and Figure~\ref{fig:problem}.





\subsection{Multi-Modal Iterative Focus Refinement}\label{sec:refinement}

We define ``multi-modal focuses'' of a CIR model as the minimal set of image segments (image tokens) and instruction keywords (text tokens) indispensable to retrieval results. Image tokens are obtained through segmentation (Segment Anything~\cite{kirillov2023segment} in this paper), while text tokens are obtained via word-level splitting. These multi-modal tokens are jointly refined through an iterative pruning process. Given an image-text query and a candidate set, \methodname~iteratively removes less-deterministic tokens while preserving those essential to maintaining the original retrieval rankings. Following the practice in~\cite{fong2017interpretable}, image tokens are pruned by zero-masking, and text tokens are pruned by replacing them with empty strings.

\begin{figure}[t]
\centering
\includegraphics[width=0.9\linewidth]{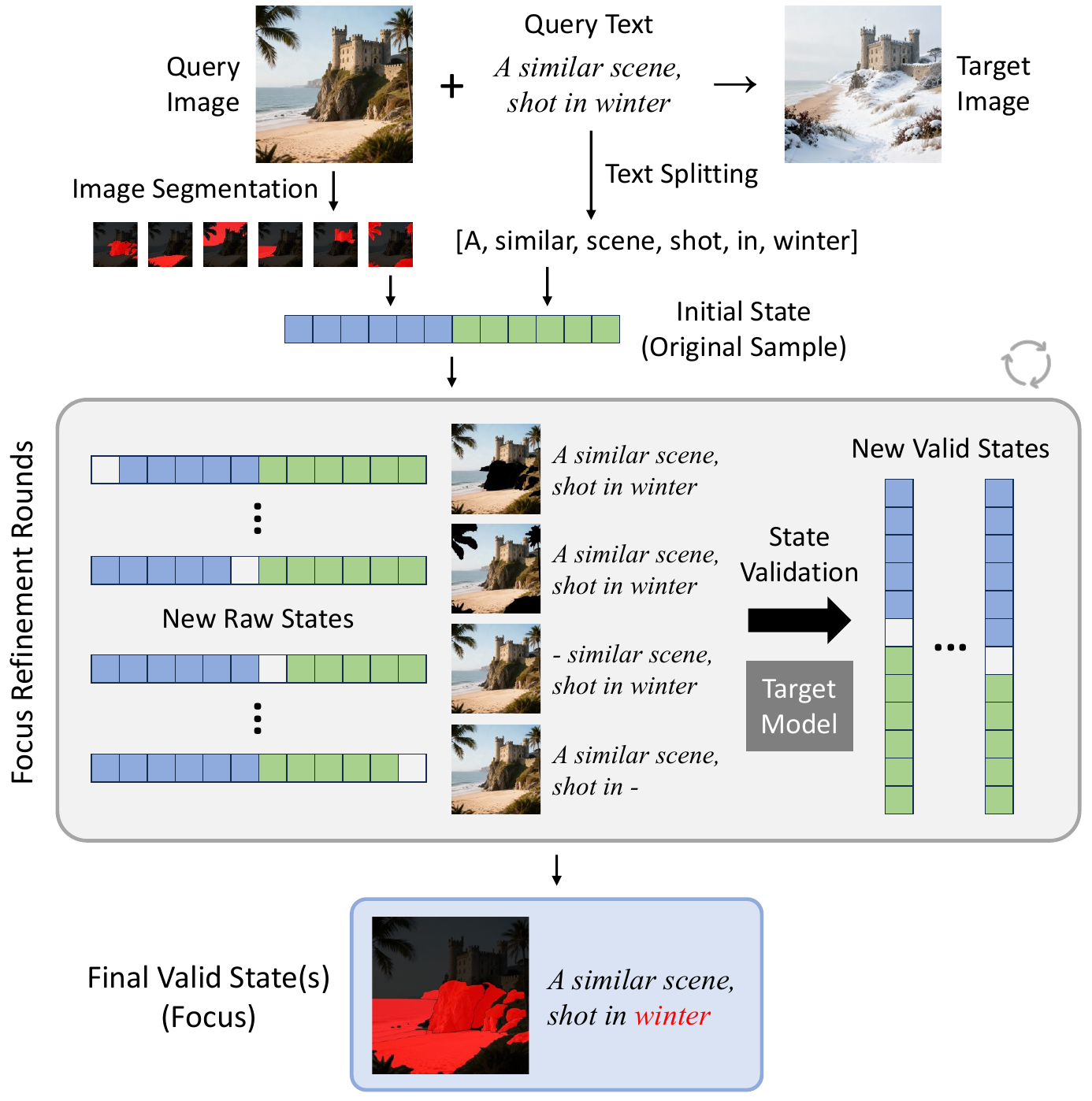}
\caption{The multi-modal iterative focus refinement process of \methodname.}
\vspace{-0.2cm}
\label{fig:refinement}
\end{figure}

For each input state $s$ with zero or more tokens pruned, we use an index set $K_s$ to store the indices of all preserved tokens of the state. The process starts with an initial state $s_0$ where all tokens are preserved and $K_{s_0}=\{1,2,\cdots,n_I+n_T\}$, given that the query image $I$ has $n_I$ tokens and the query text $T$ has $n_T$ tokens.
In each iteration, new states are derived by pruning exactly one additional token from existing states. Each new state is validated via model inference: states that preserve the original retrieval result are retained and propagated to the next iteration. The process terminates when no further valid states can be obtained. Note that multiple final states may exist for a single query, corresponding to different minimal focus configurations.
A demonstration of the process is shown in Figure~\ref{fig:refinement}. Following~\cite{focalogic}, to improve efficiency, we adopt a beam-search strategy~\cite{bisiani1992beam} and set a maximum number of new valid states per iteration (5 in this paper), which significantly accelerates refinement without degrading focus precisions.

The resulting final state set $S$ satisfies:
\begin{itemize}
\item \textbf{Validity}: All final states yield the same retrieval result as the original input: $\forall s\in S,\text{rankings}(s)=\text{rankings}(s_0)$.
\item \textbf{Minimality}: No further token removal is possible without altering the retrieval result: $\forall \hat{s}\notin S,\forall s\in S,|K_{\hat{s}}|<|K_s|\rightarrow\text{rankings}(\hat{s})\neq\text{rankings}(s_0)$.
\end{itemize}

Tokens preserved in each final state are therefore indispensable to model decisions, providing a precise and straightforward input-level interpretation of model focuses.

\subsection{Quantitatively Assessing Cross-Modal Focus Balances}

Based on the refinement results obtained in Section~\ref{sec:refinement}, we further propose a novel metric named \textit{focus balance ratio} to quantify cross-modal \textit{focus imbalances}.

For each final state $s$ and modality $M\in\{I,T\}$, \methodname~extracts its corresponding index subset $K_{s,M}$ and calculates its \textit{focus token proportion}:
\begin{equation}
p_{s,M}=\sum\alpha_M\cdot\mathrm{1}_{K_{s,M}}
\end{equation}
in which $\mathrm{1}_{K_{s,M}}$ is a binary vector with preserved indices activated, weighted by a token importance vector $\alpha_M$. $\alpha_I$ is the area proportions of the segments to the whole image, while text tokens are uniformly weighted (i.e., $\alpha_T=\frac{1}{n_T}$). Note that although initial image and text tokens may possess different semantic weights from such formulation, obtaining reliable ground-truth token weights is inherently challenging. As the weights are only applied to the preserved most crucial tokens, treating them with area-proportional (image) and uniform (text) weights is a valid approximation of the model's ``dependency'' on the respective modalities.

The \textit{focus balance ratio} of the model can then be represented by the mean \textit{relative} focus token proportion:
\begin{equation}
r_M=\frac{1}{|S|}\sum_s\frac{p_{s,M}}{\sum_{M'\in\{I,T\}}p_{s,M'}}
\end{equation}

The absolute difference between the focus balance ratios $|r_I-r_T|$ over a validation set reflects the global focus imbalances of the model: Ideally, a well-balanced CIR model should exhibit similar $r_M$s across modalities. Large gaps between $r_I$ and $r_T$ indicate the model's over-reliance on the single modality with higher $r_M$, potentially revealing shortcut behavior. Note that we do not assume a fixed oracle threshold for balance. Instead, we analyze relative differences across models and experimental settings. Regardless of the absolute values, a reduction of $|r_I-r_T|$ is highly possible to indicate an improvement of focus balances.



%% file: sections/04benchmark.tex
\section{The \dataname~Workflow and Datasets}\label{sec:benchmark}

The focus balance ratio in \methodname~primarily serves as behavior interpretations rather than direct performance metrics.

To further highlight the severity of focus imbalance and to explore potential solutions, we propose a novel dataset augmentation workflow for CIR. This workflow produces two datasets: a benchmark dataset designed to quantitatively evaluate existing CIR models on focus-challenging cases, and a finetuning dataset intended to encourage more balanced cross-modal focuses during training. Both datasets are constructed under a unified paradigm stated below.

\begin{figure}[t]
\centering
\includegraphics[width=\linewidth]{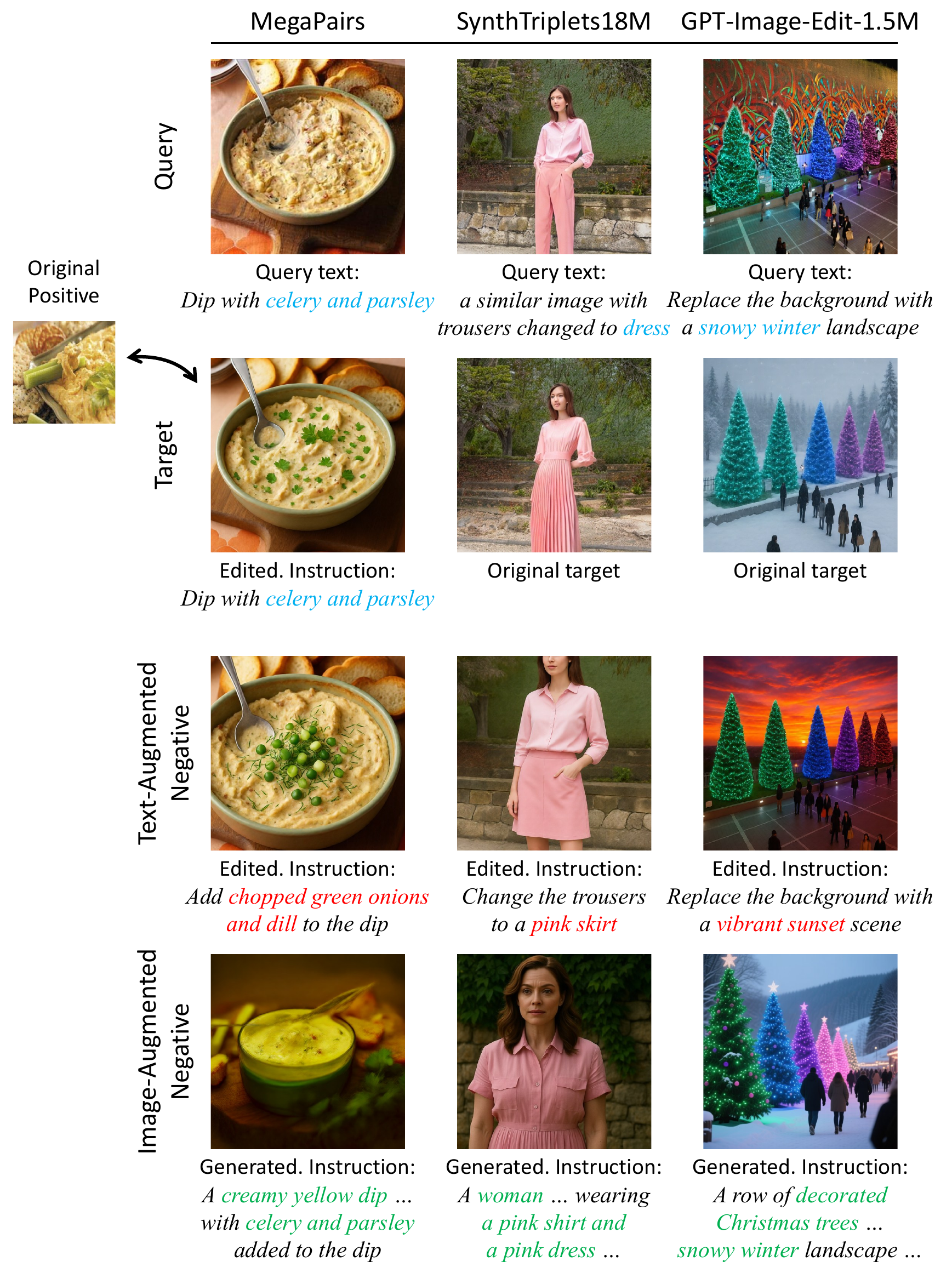}
\caption{Examples of the data constructed by the \dataname~workflow. For the real-life dataset MegaPairs, we synthesize a positive for each triplet, and regard the original positive a special candidate.}
\vspace{-0.2cm}
\label{fig:data}
\end{figure}

\subsection{Data Construction}\label{sec:data}

The core idea of the \dataname~workflow is to expose focus imbalances through crafted hard negatives. If the model overfocuses on the image modality, its hardest confusions are candidates visually similar to the query image but semantically inconsistent with the query text. Conversely, when a model overfocuses on text, its hard negatives are semantically aligned with the query text while exhibiting visual discrepancies. In both cases, neglecting one modality leads to incorrect retrieval results.

The workflow incorporates a VLM (Qwen3-VL-30B-A3B-Instruct~\cite{yang2025qwen3} in this paper), an image editing model (Qwen-Image-Edit~\cite{wu2025qwen} in this paper), and an image generation model (Qwen-Image~\cite{wu2025qwen} in this paper), and operates on existing CIR triplets. Starting from a triplet, the workflow constructs a focus-challenging sample with the following components:
\begin{itemize}
\item \textbf{Query}: The original query from the triplet.
\item \textbf{Text-augmented negatives}: Leveraging the VLM, the workflow alters the semantics of the query text to produce multiple modified negative texts. Then, the workflow conducts image editing on the query image with the modified negative texts to yield multiple negative images that remain visually similar to the query image while violating the textual intent, highlighting overfocuses on the image modality.
\item \textbf{Image-augmented negatives}: Leveraging the VLM, the workflow generates a comprehensive text description that integrates the semantics of the query. Then, the workflow conducts image generation with the description to obtain multiple negative images that are semantically aligned with the query text while having visual discrepancies from the query image, highlighting overfocuses on the text modality.
\item \textbf{Identity negatives}: The original query image itself is included as a hard negative, given that there is no query text for identity retrieval.
\item \textbf{Positive (Target)}: Triplets in most existing real-world CIR datasets are constructed via similarity-driven image pairing and query text generation~\cite{zhou2025megapairs}. This results in relatively low data quality, as a large number of targets are only loosely consistent with their queries. To address this issue, the workflow treats the original positive from such datasets as a special candidate rather than the truth target, and synthesizes a strictly-consistent positive image based on the query. For datasets built under editing-driven triplet construction paradigms~\cite{wang2025gpt,gu2023compodiff}, the target is intrinsically consistent with the query, and the original positive is retained.
\end{itemize}

Since real-world datasets lack such specific counterfactual negatives required to isolate focus imbalances, the synthetic approach is the only way to quantitatively measure this specific model behavior.

The above construction strategy is applied to both the benchmark and the finetuning dataset. We offer detailed quality analyses of them in Appendix~\ref{app:data}. Specifically, for the \dataname~benchmark, its statistics and comparisons with existing CIR benchmarks are reported in Table~\ref{tab:benchmark}. Representative examples of the constructed data are shown in Figure~\ref{fig:data}.

\subsection{Quantitative Metrics in the Benchmark}

The proposed benchmark is designed to focus only on in-sample subset retrieval performance, as existing benchmarks~\cite{liu2021image,baldrati2023zero,vaze2023genecis,wu2021fashion} already provide comprehensive evaluations on models' global retrieval capabilities. Specifically, the benchmark employs two quantitative metrics:
\begin{itemize}
\item \textbf{Subset recall ($R_s$)}: Recall computed over the candidate set of each sample, consisting of the positive and the multiple types of hard negatives stated in Section~\ref{sec:data}. This metric measures model performance under focus-challenging conditions.
\item \textbf{Focus imbalance ($|r_I-r_T|$)}: This metric directly quantifies models' cross-modal focus imbalances.
\end{itemize}
Together, these metrics characterize CIR models from two complementary perspectives: Subset recall reflects task performance under challenging cases, while focus imbalance captures underlying focus-related model behavior.

\input{tables/benchmark}

%% file: tables/benchmark.tex
\begin{table}[t]
\centering
\small
\caption{Statistical comparisons between the \dataname~and the standard benchmarks. Note that our proposed benchmark can be easily expanded based on existing triplets.}
\begin{tabular}{cccc}
\toprule
Benchmark & Queries & Candidates & Candidates (Local) \\
\midrule
CIRR & 4180 & 2265 & 5.97 \\
FashionIQ & 12032 & 6003 & N/A \\
GeneCIS & 3920 & 3030 & 15.0 \\
\dataname & 1000 & 3844 & 5.0 \\
\bottomrule
\end{tabular}
\vspace{-0.2cm}
\label{tab:benchmark}
\end{table}

%% file: sections/05experiments.tex
\section{Experiments and Discussions}

\subsection{Experimental Setup}

We construct the \dataname~benchmark and finetuning dataset based on three large-scale CIR datasets: MegaPairs~\cite{zhou2025megapairs}, SynthTriplets18M~\cite{gu2023compodiff}, and GPT-Image-Edit-1.5M~\cite{wang2025gpt}. MegaPairs consists of real-world images, while the other two datasets are synthesized via image editing. Note that the text instructions in GPT-Image-Edit-1.5M are notably complex and frequently involve multiple editing requirements. Therefore, we decompose them into multiple simpler units to better align with the text distribution in standard CIR settings. To reflect realistic retrieval scenarios, we sample data from the three datasets with a ratio of 3:1:1. In each constructed sample, the numbers of positives, text-augmented negatives, and image-augmented negatives are all set to 1. In addition to the proposed hard-case benchmark, we also conduct extensive model performance evaluations on three widely-used standard benchmarks - CIRR~\cite{liu2021image}, FashionIQ~\cite{wu2021fashion}, and GeneCIS~\cite{vaze2023genecis} - for general retrieval performance analysis.

The tested models include the CLIP-based CLIP4CIR~\cite{baldrati2023composed}, SEARLE~\cite{baldrati2023zero}, and BGE~\cite{zhou2025megapairs} families, and the VLM-based GME~\cite{zhang2024gme} family, RzenEmbed~\cite{jian2025rzenembed}, and MM-Embed~\cite{lin2024mm}. All models go through the identical evaluation and finetuning process.

\subsection{Pretrained Model Performance on the \dataname~Benchmark}\label{sec:pretrained}

Model performance on the \dataname~and the standard benchmarks are summarized in Table~\ref{tab:pretrained}. Results show that the tested models achieve comparable retrieval recalls when retrieving from large global candidate pools (with K-magnitude numbers of samples) in standard benchmarks and from small local candidate pools (with 5 samples in average) in the \dataname~benchmark. This observation highlights the difficulty of the proposed benchmark and suggests the presence of significant focus imbalances in existing models. Moreover, models exhibit inconsistent trends between common-case performance and focus balance, indicating that current benchmarks are insufficient for comprehensively evaluating model performance and behavior. Additional experimental results and qualitative focus visualizations are provided in Appendix~\ref{app:focus_vis}.

\input{tables/pretrained}

\subsection{Finetuning Models on the \dataname~Dataset}\label{sec:finetune}

\input{tables/finetune_recall}

We finetune the models for 1 epoch using the constructed dataset and evaluate their zero-shot (as the constructed data has no overlap with existing benchmarks) performance. We finetune the last few layers of the CLIP-based models, and apply LoRA~\cite{hu2021lora} to the VLM-based models. Detailed finetuning configurations are provided in Appendix~\ref{app:exp_details}. Unless otherwise specified, augmentation is applied to 50\% of the finetuning samples, with the remaining samples consisting of standard CIR triplets. Results under more experimental settings are reported in Section~\ref{sec:ablation} and Appendix~\ref{app:finetune_all}.

Table~\ref{tab:finetune_recall} shows the finetuning results of the VLM-based models on both the standard and the \dataname~benchmarks. Results of the CLIP-based models are included in Appendix~\ref{app:finetune_clip}. Across multiple experimental settings, the models consistently improve after finetuning, demonstrating the effectiveness of the proposed dataset in both common-case and hard-case scenarios. Notably, recall gains on hard-case performance (\dataname~Rs@1) are substantially larger than those on standard benchmarks, indicating that the proposed dataset primarily improves robustness under focus-challenging conditions rather than general abilities. Model improvements in focus balances are presented in Table~\ref{tab:finetune_focus}. The results directly confirm that the proposed dataset effectively encourages more balanced cross-modal focuses.

\input{tables/finetune_focus}

\input{tables/finetune_recall_zeroshot}

\input{tables/finetune_recall_neg_ratio_all}

Specifically, improvements on CIRR Rs@1 (with candidates semantically closer than other settings) are generally larger than gains on other metrics, further demonstrating the effectiveness of the proposed dataset in improving models' zero-shot hard-case performance. More related analyses are presented in Appendix~\ref{app:zeroshot_analysis}. To further validate its generalization capabilities, we apply the identical augmentation strategy to CIRR to construct a distribution-shifted zero-shot hard-case benchmark (\methodname-CIRR). As shown in Table~\ref{tab:finetune_recall_zeroshot}, the models finetuned with the proposed dataset consistently outperform their pretrained counterparts, confirming that the observed improvements are stable and transferable.

\subsection{Ablation Studies}\label{sec:ablation}

As stated in Section~\ref{sec:finetune}, the proposed augmentation strategy can be applied to only a subset of the finetuning data, and the proportion of augmented samples is adjustable. In this section, we analyze the impacts of data scales and negative ratios on finetuned model performance.

Table~\ref{tab:finetune_recall_neg_ratio_all} reports results under varying data scales and negative ratios. When the negative ratio is set to 0\%, the data contains no \methodname~augmentation and serves as a baseline. Compared to this baseline, \methodname-augmented data consistently yields performance improvements across both standard and hard-case benchmarks, demonstrating the superiority of the proposed workflow. Specifically, model performance on the \dataname~benchmark improves monotonically with negative ratios and data scales, highlighting the effectiveness of more difficult and more diverse negatives in mitigating focus imbalances. Additional analyses on focus imbalances under different configurations are provided in Appendix~\ref{app:focus_all}.

%% file: tables/pretrained.tex
\begin{table}[t]
\centering
\small
\caption{Pretrained performance of the tested CIR models on the standard and the \dataname~benchmarks. The ``Average'' column shows the average value of CIRR R@1, CIRR Rs@1, FashionIQ R@1, and GeneCIS Rs@1.}
\begin{tabular}{cccc}
\toprule
\multirow{2}{*}{Model} & \multirow{2}{*}{Average} & \dataname & Focus \\
& & Rs@1 & Imbalance \\
\midrule
\multicolumn{4}{c}{CLIP-based Models} \\
\midrule
CLIP4CIR-RN50 & 24.9 & 33.4 & 0.53 \\
CLIP4CIR-RN50x4 & 26.0 & 31.2 & 0.65 \\
SEARLE-base & 23.6 & 29.3 & 0.27 \\
SEARLE-large & 24.7 & 30.2 & 0.70 \\
BGE-base & 36.5 & 41.6 & 0.10 \\
BGE-large & 38.3 & 38.8 & 0.02 \\
\midrule
\multicolumn{4}{c}{VLM-based Models} \\
\midrule
GME-2B & 33.8 & 29.7 & 0.54 \\
GME-7B & 36.1 & 32.9 & 0.55 \\
RzenEmbed-7B & 37.6 & 39.3 & 0.43 \\
MM-Embed-7B & 27.2 & 26.7 & 0.60 \\
\bottomrule
\end{tabular}
\vspace{-0.2cm}
\label{tab:pretrained}
\end{table}

%% file: tables/finetune_recall.tex
\begin{table*}[t]
\centering
\small
\caption{Finetuned performance of the VLM-based CIR models trained with different data scales and LoRA ranks on the standard and the \dataname~benchmarks. Higher is better.}
\begin{tabular}{c|ccc|ccc}
\toprule
\multicolumn{7}{c}{Metrics: CIRR R@1 / CIRR Rs@1 / FashionIQ R@10 / GeneCIS Rs@1 $\ ||\ $ Average (Gain) $\ ||\ $ \dataname~Rs@1 (Gain)} \\
\midrule
Data / Rank & \multicolumn{3}{c|}{GME-2B} & \multicolumn{3}{c}{GME-7B} \\
\midrule
Pretrained & 32.8 / 41.5 / 39.8 / 21.1 & 33.8 & 29.7 & 35.2 / 44.3 / 43.3 / 21.7 & 36.1 & 32.9 \\
\midrule
5K / 2 & 36.8 / 48.3 / 40.0 / 20.9 & 36.5 (\textcolor{green}{+2.7}) & 37.6 (\textcolor{green}{+7.9}) & 41.1 / 51.1 / 43.2 / 21.7 & 39.3 (\textcolor{green}{+3.1}) & 42.2 (\textcolor{green}{+9.3}) \\
10K / 2 & 37.1 / 48.4 / 39.7 / 21.3 & 36.6 (\textcolor{green}{+2.8}) & 37.7 (\textcolor{green}{+8.0}) & 44.1 / 55.3 / 43.2 / 21.8 & 41.1 (\textcolor{green}{+5.0}) & 42.7 (\textcolor{green}{+9.8}) \\
20K / 2 & 35.3 / 45.6 / 39.9 / 20.5 & 35.3 (\textcolor{green}{+1.5}) & 36.6 (\textcolor{green}{+6.9}) & 43.5 / 54.6 / 43.2 / 21.7 & 40.7 (\textcolor{green}{+4.6}) & 44.2 (\textcolor{green}{+11.3}) \\
50K / 2 & 36.8 / 47.7 / 40.1 / 20.8 & 36.4 (\textcolor{green}{+2.6}) & 37.9 (\textcolor{green}{+8.2}) & 44.5 / 56.2 / 43.0 / 21.4 & 41.3 (\textcolor{green}{+5.1}) & 46.6 (\textcolor{green}{+13.7}) \\
\midrule
5K / 4 & 35.5 / 45.7 / 40.1 / 21.3 & 35.6 (\textcolor{green}{+1.9}) & 35.8 (\textcolor{green}{+6.1}) & 43.6 / 54.8 / 43.4 / 22.0 & 40.9 (\textcolor{green}{+4.8}) & 43.1 (\textcolor{green}{+10.2}) \\
10K / 4 & 35.7 / 45.7 / 39.8 / 20.4 & 35.4 (\textcolor{green}{+1.6}) & 36.1 (\textcolor{green}{+6.4}) & 42.9 / 53.8 / 43.2 / 21.5 & 40.3 (\textcolor{green}{+4.2}) & 43.4 (\textcolor{green}{+10.5}) \\
20K / 4 & 37.1 / 48.2 / 40.1 / 21.1 & 36.6 (\textcolor{green}{+2.8}) & 40.7 (\textcolor{green}{+11.0}) & 42.5 / 52.9 / 43.2 / 21.9 & 40.1 (\textcolor{green}{+4.0}) & 45.0 (\textcolor{green}{+12.1}) \\
50K / 4 & 36.5 / 47.8 / 39.6 / 20.5 & 36.1 (\textcolor{green}{+2.3}) & 38.8 (\textcolor{green}{+9.1}) & 46.4 / 59.2 / 43.3 / 21.9 & 42.7 (\textcolor{green}{+6.6}) & 47.3 (\textcolor{green}{+14.4}) \\
\midrule
Data / Rank & \multicolumn{3}{c|}{RzenEmbed-7B} & \multicolumn{3}{c}{MM-Embed-7B} \\
\midrule
Pretrained & 40.1 / 47.5 / 41.2 / 21.4 & 37.6 & 39.3 & 24.5 / 27.2 / 38.2 / 18.9 & 27.2 & 26.7 \\
\midrule
5K / 2 & 41.5 / 49.8 / 41.3 / 22.1 & 38.6 (\textcolor{green}{+1.1}) & 42.2 (\textcolor{green}{+2.9}) & 26.6 / 30.0 / 38.3 / 19.1 & 28.5 (\textcolor{green}{+1.3}) & 30.9 (\textcolor{green}{+4.2}) \\
10K / 2 & 43.7 / 53.5 / 40.7 / 23.6 & 40.4 (\textcolor{green}{+2.8}) & 44.7 (\textcolor{green}{+5.4}) & 28.3 / 32.2 / 38.6 / 19.1 & 29.6 (\textcolor{green}{+2.4}) & 34.2 (\textcolor{green}{+7.5}) \\
20K / 2 & 45.5 / 55.9 / 40.8 / 23.0 & 41.3 (\textcolor{green}{+3.7}) & 44.1 (\textcolor{green}{+4.8}) & 30.2 / 35.0 / 39.0 / 20.7 & 31.2 (\textcolor{green}{+4.0}) & 35.0 (\textcolor{green}{+8.3}) \\
50K / 2 & 44.6 / 54.8 / 40.6 / 22.9 & 40.7 (\textcolor{green}{+3.2}) & 45.4 (\textcolor{green}{+6.1}) & 31.3 / 36.8 / 38.9 / 20.6 & 31.9 (\textcolor{green}{+4.7}) & 38.5 (\textcolor{green}{+11.8}) \\
\midrule
5K / 4 & 42.2 / 51.1 / 41.5 / 22.1 & 39.2 (\textcolor{green}{+1.7}) & 42.4 (\textcolor{green}{+3.1}) & 24.5 / 27.4 / 38.8 / 18.2 & 27.2 (+0.0) & 30.2 (\textcolor{green}{+3.5}) \\
10K / 4 & 41.6 / 49.8 / 41.6 / 22.3 & 38.9 (\textcolor{green}{+1.3}) & 42.5 (\textcolor{green}{+3.2}) & 27.4 / 30.8 / 38.4 / 20.6 & 29.3 (\textcolor{green}{+2.1}) & 32.8 (\textcolor{green}{+6.1}) \\
20K / 4 & 40.5 / 47.4 / 41.4 / 21.5 & 37.7 (\textcolor{green}{+0.1}) & 42.8 (\textcolor{green}{+3.5}) & 31.3 / 36.6 / 38.4 / 21.4 & 31.9 (\textcolor{green}{+4.7}) & 35.7 (\textcolor{green}{+9.0}) \\
50K / 4 & 43.0 / 50.8 / 40.9 / 22.4 & 39.3 (\textcolor{green}{+1.7}) & 45.5 (\textcolor{green}{+6.2}) & 26.3 / 29.2 / 37.8 / 20.1 & 28.4 (\textcolor{green}{+1.2}) & 35.7 (\textcolor{green}{+9.0}) \\
\bottomrule
\end{tabular}
\vspace{-0.2cm}
\label{tab:finetune_recall}
\end{table*}

%% file: tables/finetune_focus.tex
\begin{table*}[t]
\centering
\small
\caption{Finetuned focus imbalances of the VLM-based CIR models trained with different data scales and LoRA ranks. Lower is better.}
\begin{tabular}{c|cccc|cccc}
\toprule
\multicolumn{9}{c}{Metrics: Focus Imbalance (Reduction)} \\
\midrule
Data & 5K & 10K & 20K & 50K & 5K & 10K & 20K & 50K \\
\midrule
Rank & \multicolumn{4}{c|}{GME-2B (Pretrained: 0.54)} & \multicolumn{4}{c}{GME-7B (Pretrained: 0.55)} \\
\midrule
2 & 0.51 (\textcolor{green}{-0.03}) & 0.58 (\textcolor{red}{+0.04}) & 0.21 (\textcolor{green}{-0.33}) & 0.28 (\textcolor{green}{-0.26}) & 0.54 (\textcolor{green}{-0.01}) & 0.54 (\textcolor{green}{-0.02}) & 0.23 (\textcolor{green}{-0.33}) & 0.13 (\textcolor{green}{-0.42}) \\
4 & 0.56 (\textcolor{red}{+0.02}) & 0.40 (\textcolor{green}{-0.14}) & 0.17 (\textcolor{green}{-0.38}) & 0.32 (\textcolor{green}{-0.22}) & 0.54 (\textcolor{green}{-0.01}) & 0.54 (\textcolor{green}{-0.02}) & 0.23 (\textcolor{green}{-0.33}) & 0.13 (\textcolor{green}{-0.42}) \\
\midrule
Rank & \multicolumn{4}{c|}{RzenEmbed-7B (Pretrained: 0.43)} & \multicolumn{4}{c}{MM-Embed-7B (Pretrained: 0.60)} \\
\midrule
2 & 0.44 (\textcolor{red}{+0.01}) & 0.23 (\textcolor{green}{-0.21}) & 0.10 (\textcolor{green}{-0.33}) & 0.11 (\textcolor{green}{-0.32}) & 0.56 (\textcolor{green}{-0.04}) & 0.05 (\textcolor{green}{-0.55}) & 0.06 (\textcolor{green}{-0.54}) & 0.16 (\textcolor{green}{-0.44}) \\
4 & 0.44 (\textcolor{red}{+0.01}) & 0.08 (\textcolor{green}{-0.35}) & 0.21 (\textcolor{green}{-0.22}) & 0.04 (\textcolor{green}{-0.39}) & 0.67 (\textcolor{red}{+0.07}) & 0.17 (\textcolor{green}{-0.43}) & 0.05 (\textcolor{green}{-0.55}) & 0.14 (\textcolor{green}{-0.45}) \\
\bottomrule
\end{tabular}
\vspace{-0.2cm}
\label{tab:finetune_focus}
\end{table*}

%% file: tables/finetune_recall_zeroshot.tex
\begin{table*}[!t]
\centering
\small
\caption{Finetuned zero-shot performance of the VLM-based CIR models trained with different data scales and LoRA ranks on the \methodname-augmented CIRR benchmark. Higher is better.}
\begin{tabular}{c|cccc|cccc}
\toprule
\multicolumn{9}{c}{Metrics: \methodname-CIRR Rs@1 (Gain)} \\
\midrule
Data & 5K & 10K & 20K & 50K & 5K & 10K & 20K & 50K \\
\midrule
Rank & \multicolumn{4}{c|}{GME-2B (Pretrained: 28.5)} & \multicolumn{4}{c}{GME-7B (Pretrained: 32.9)} \\
\midrule
2 & 29.8 (\textcolor{green}{+1.3}) & 29.6 (\textcolor{green}{+1.1}) & 28.0 (\textcolor{red}{-0.5}) & 30.2 (\textcolor{green}{+1.7}) & 27.4 (\textcolor{green}{+2.4}) & 27.7 (\textcolor{green}{+2.7}) & 27.3 (\textcolor{green}{+2.3}) & 29.0 (\textcolor{green}{+4.0}) \\
4 & 29.2 (\textcolor{green}{+0.7}) & 28.7 (\textcolor{green}{+0.2}) & 29.7 (\textcolor{green}{+1.2}) & 30.0 (\textcolor{green}{+1.5}) & 27.3 (\textcolor{green}{+2.3}) & 28.7 (\textcolor{green}{+3.7}) & 29.4 (\textcolor{green}{+4.4}) & 30.1 (\textcolor{green}{+5.1}) \\
\midrule
Rank & \multicolumn{4}{c|}{RzenEmbed-7B (Pretrained: 30.1)} & \multicolumn{4}{c}{MM-Embed-7B (Pretrained: 26.7)} \\
\midrule
2 & 30.5 (\textcolor{green}{+0.4}) & 31.5 (\textcolor{green}{+1.4}) & 30.4 (\textcolor{green}{+0.3}) & 33.2 (\textcolor{green}{+3.1}) & 29.4 (\textcolor{green}{+1.9}) & 31.5 (\textcolor{green}{+4.0}) & 29.5 (\textcolor{green}{+2.0}) & 33.3 (\textcolor{green}{+5.8}) \\
4 & 30.8 (\textcolor{green}{+0.7}) & 32.0 (\textcolor{green}{+1.9}) & 32.2 (\textcolor{green}{+2.1}) & 34.0 (\textcolor{green}{+3.9}) & 30.7 (\textcolor{green}{+3.2}) & 30.8 (\textcolor{green}{+3.3}) & 31.9 (\textcolor{green}{+4.4}) & 30.9 (\textcolor{green}{+3.4}) \\
\bottomrule
\end{tabular}
\vspace{-0.2cm}
\label{tab:finetune_recall_zeroshot}
\end{table*}

%% file: tables/finetune_recall_neg_ratio_all.tex
\begin{table*}[h]
\centering
\small
\caption{Finetuned (with 2-rank LoRA) performance of the VLM-based CIR models trained with different data scales and negative ratios on the \dataname~benchmark. Higher is better.}
\begin{tabular}{c|cccc|cccc}
\toprule
\multicolumn{9}{c}{Metrics: Average (Gain) / \dataname~Rs@1 (Gain)} \\
\midrule
Neg & 5K & 10K & 20K & 50K & 5K & 10K & 20K & 50K \\
\cmidrule{2-9}
Ratio & \multicolumn{4}{c|}{GME-2B (Pretrained: 33.8 / 29.7)} & \multicolumn{4}{c}{GME-7B (Pretrained: 36.1 / 32.9)} \\
\midrule
0\% & 35.7 / 33.6 & 36.0 / 33.1 & 35.3 / 34.5 & 35.6 / 33.3 & 38.8 / 38.6 & 39.6 / 38.6 & 39.2 / 39.3 & 39.4 / 39.1 \\
\midrule
\multirow{2}{*}{25\%} & 35.9 (\textcolor{green}{+0.2}) & 36.2 (\textcolor{green}{+0.2}) & 37.2 (\textcolor{green}{+1.8}) & 36.2 (\textcolor{green}{+0.6}) & 40.4 (\textcolor{green}{+1.6}) & 39.3 (\textcolor{red}{-0.2}) & 38.5 (\textcolor{red}{-0.7}) & 39.6 (\textcolor{green}{+0.1}) \\
& 34.6 (\textcolor{green}{+1.0}) & 36.2 (\textcolor{green}{+3.1}) & 37.2 (\textcolor{green}{+2.7}) & 37.8 (\textcolor{green}{+4.5}) & 40.5 (\textcolor{green}{+1.9}) & 41.5 (\textcolor{green}{+2.9}) & 40.3 (\textcolor{green}{+1.0}) & 41.1 (\textcolor{green}{+2.0}) \\
\midrule
\multirow{2}{*}{50\%} & 36.5 (\textcolor{green}{+0.8}) & 36.6 (\textcolor{green}{+0.6}) & 35.3 (+0.0) & 36.4 (\textcolor{green}{+0.7}) & 39.3 (\textcolor{green}{+0.5}) & 41.1 (\textcolor{green}{+1.6}) & 40.7 (\textcolor{green}{+1.6}) & 41.3 (\textcolor{green}{+1.8}) \\
& 37.6 (\textcolor{green}{+4.0}) & 37.7 (\textcolor{green}{+4.6}) & 36.6 (\textcolor{green}{+2.1}) & 37.9 (\textcolor{green}{+4.6}) & 42.2 (\textcolor{green}{+3.6}) & 42.7 (\textcolor{green}{+4.1}) & 44.2 (\textcolor{green}{+4.9}) & 46.6 (\textcolor{green}{+7.5}) \\
\midrule
\multirow{2}{*}{75\%} & 36.7 (\textcolor{green}{+1.0}) & 37.0 (\textcolor{green}{+1.0}) & 35.4 (+0.0) & 35.0 (\textcolor{red}{-0.6}) & 41.1 (\textcolor{green}{+2.3}) & 41.3 (\textcolor{green}{+1.7}) & 41.2 (\textcolor{green}{+2.1}) & 39.9 (\textcolor{green}{+0.5}) \\
& 39.4 (\textcolor{green}{+5.8}) & 39.5 (\textcolor{green}{+6.4}) & 40.7 (\textcolor{green}{+6.2}) & 42.0 (\textcolor{green}{+8.7}) & 44.9 (\textcolor{green}{+6.3}) & 45.7 (\textcolor{green}{+7.1}) & 47.3 (\textcolor{green}{+8.0}) & 47.7 (\textcolor{green}{+8.6}) \\
\midrule
\multirow{2}{*}{100\%} & 36.6 (\textcolor{green}{+0.9}) & 36.4 (\textcolor{green}{+0.4}) & 34.4 (\textcolor{red}{-0.9}) & 36.5 (\textcolor{green}{+0.9}) & 40.2 (\textcolor{green}{+1.4}) & 41.7 (\textcolor{green}{+2.2}) & 39.1 (+0.0) & 39.8 (\textcolor{green}{+0.4}) \\
& 40.7 (\textcolor{green}{+7.1}) & 42.6 (\textcolor{green}{+9.5}) & 41.1 (\textcolor{green}{+6.6}) & 45.5 (\textcolor{green}{+12.2}) & 45.0 (\textcolor{green}{+6.4}) & 49.3 (\textcolor{green}{+10.7}) & 48.7 (\textcolor{green}{+9.4}) & 50.4 (\textcolor{green}{+11.3}) \\
\midrule
& \multicolumn{4}{c|}{RzenEmbed-7B (Pretrained: 37.6 / 39.3)} & \multicolumn{4}{c}{MM-Embed-7B (Pretrained: 27.2 / 26.7)} \\
\midrule
0\% & 37.3 / 38.9 & 36.5 / 37.7 & 36.8 / 37.8 & 36.7 / 38.3 & 24.5 / 25.4 & 24.6 / 25.7 & 26.1 / 26.5 & 21.9 / 24.0 \\
\midrule
\multirow{2}{*}{25\%} & 38.4 (\textcolor{green}{+1.0}) & 37.1 (\textcolor{green}{+0.6}) & 38.5 (\textcolor{green}{+1.6}) & 38.8 (\textcolor{green}{+2.2}) & 26.7 (\textcolor{green}{+2.2}) & 27.3 (\textcolor{green}{+2.7}) & 28.8 (\textcolor{green}{+2.7}) & 27.6 (\textcolor{green}{+5.7}) \\
& 41.8 (\textcolor{green}{+2.9}) & 41.2 (\textcolor{green}{+3.5}) & 41.3 (\textcolor{green}{+3.5}) & 41.3 (\textcolor{green}{+3.0}) & 29.1 (\textcolor{green}{+3.7}) & 29.3 (\textcolor{green}{+3.6}) & 31.5 (\textcolor{green}{+5.0}) & 32.8 (\textcolor{green}{+8.8}) \\
\midrule
\multirow{2}{*}{50\%} & 38.6 (\textcolor{green}{+1.3}) & 40.4 (\textcolor{green}{+3.9}) & 41.3 (\textcolor{green}{+4.4}) & 40.7 (\textcolor{green}{+4.1}) & 28.5 (\textcolor{green}{+4.0}) & 29.6 (\textcolor{green}{+5.0}) & 31.2 (\textcolor{green}{+5.1}) & 31.9 (\textcolor{green}{+10.0}) \\
& 42.2 (\textcolor{green}{+3.3}) & 44.7 (\textcolor{green}{+7.0}) & 44.1 (\textcolor{green}{+6.3}) & 45.4 (\textcolor{green}{+7.1}) & 30.9 (\textcolor{green}{+5.5}) & 34.2 (\textcolor{green}{+8.5}) & 35.0 (\textcolor{green}{+8.5}) & 38.5 (\textcolor{green}{+14.5}) \\
\midrule
\multirow{2}{*}{75\%} & 40.0 (\textcolor{green}{+2.7}) & 39.4 (\textcolor{green}{+2.9}) & 38.2 (\textcolor{green}{+1.4}) & 39.7 (\textcolor{green}{+3.0}) & 34.1 (\textcolor{green}{+9.5}) & 32.4 (\textcolor{green}{+7.8}) & 32.0 (\textcolor{green}{+5.8}) & 34.8 (\textcolor{green}{+12.9}) \\
& 42.6 (\textcolor{green}{+3.7}) & 45.9 (\textcolor{green}{+8.2}) & 47.5 (\textcolor{green}{+9.7}) & 48.5 (\textcolor{green}{+10.2}) & 41.0 (\textcolor{green}{+15.6}) & 39.4 (\textcolor{green}{+13.7}) & 41.2 (\textcolor{green}{+14.7}) & 41.1 (\textcolor{green}{+17.1}) \\
\midrule
\multirow{2}{*}{100\%} & 38.9 (\textcolor{green}{+1.6}) & 40.1 (\textcolor{green}{+3.6}) & 39.2 (\textcolor{green}{+2.3}) & 41.4 (\textcolor{green}{+4.7}) & 29.4 (\textcolor{green}{+4.8}) & 33.2 (\textcolor{green}{+8.6}) & 33.0 (\textcolor{green}{+6.9}) & 32.8 (\textcolor{green}{+10.9}) \\
& 45.7 (\textcolor{green}{+6.8}) & 47.8 (\textcolor{green}{+10.1}) & 48.2 (\textcolor{green}{+10.4}) & 50.1 (\textcolor{green}{+11.8}) & 34.7 (\textcolor{green}{+9.3}) & 40.3 (\textcolor{green}{+14.6}) & 43.0 (\textcolor{green}{+16.5}) & 45.4 (\textcolor{green}{+21.4}) \\
\bottomrule
\end{tabular}
\vspace{-0.2cm}
\label{tab:finetune_recall_neg_ratio_all}
\end{table*}

%% file: sections/06limitations.tex
\section{Discussions and Limitations}

Despite the effectiveness of \methodname, this work has several limitations worth discussion.

First, the quantitative metrics in the \dataname~benchmark rely on heuristic token weighting schemes, such as region-area-based weights for image tokens and uniform weights for text tokens. Although these choices are intuitive and yield consistent trends across a wide range of samples and experimental settings, they may introduce biases in specific cases and do not explicitly model fine-grained semantic dependencies between modalities. Importantly, our analysis does not depend on absolute metric values; instead, we focus on relative comparisons across models and training settings, which mitigates the impacts of such biases in practice.

Second, the focus refinement process in \methodname~requires multiple model inferences. This results in relatively high time complexity of the method. As a result, the method is not optimized for large-scale or real-time deployment, and is better suited for evaluation and diagnostic purposes. Nevertheless, the process remains tractable for benchmark construction and model analyses. Empirical runtime statistics are provided in Appendix~\ref{app:computation}.

Overall, these limitations point to promising directions for future work, including more principled weighting strategies and more efficient focus identification mechanisms, while not affecting the practical insights and empirical findings presented in this work.

%% file: sections/07conclusion.tex
\section{Conclusions}

In this paper, we investigate the problem of \textit{focus imbalances} in composed image retrieval models, where models over-rely on a single input modality while neglecting complementary information from the other, leading to degraded performance on hard negatives. To study this phenomenon, we first propose a multi-modal model focus interpretation method named \methodname~that enables direct analyses of models' focus-related behavior and validates the prevalence and severity of the problem. Building on the analyses, we further propose the \dataname~workflow for data augmentation. Leveraging the workflow, we construct a dedicated benchmark for quantitatively evaluating focus imbalances, and a finetuning dataset to encourage more balanced cross-modal focuses. Experiments validate the effectiveness of \methodname~and the constructed datasets in improving models' hard-case performance while maintaining their capabilities on standard benchmarks. This work provides a complementary evaluation perspective in CIR, and offers a new dimension for analyzing and improving the precision and robustness of CIR models.

%% file: sections/impact.tex
\section*{Impact Statement}

The primary goal of this work is to improve robustness and interpretability of machine learning models in the field of composed image retrieval. The techniques introduced in this work have positive impacts on practical applications such as recommendation systems and human-machine interaction. This work introduces no new ethical concerns beyond those commonly raised by existing vision-language models and AI-generated content, and does not need specific highlights or additional discussions in this paper.

%% file: sections/appendix.tex
\section{Significance of Focus Balancing in Improving Model Performance}\label{app:focus_analysis}

In this section, we provide a more detailed discussion of focus balancing and its impact on CIR performance by addressing the following three questions:

\textbf{Question \#1: From which perspective does focus balancing contribute to more stable CIR performance?}

As illustrated in Figure~\ref{fig:problem} in the main paper, models with imbalanced focuses tend to fail on hard cases where the negative candidate pool contains the semantics from both input modalities. In such cases, correct retrieval requires the model to jointly reason over visual and textual information, making balanced focus a necessary condition for stable CIR performance.

In contrast, common-case scenarios are often dominated by large semantic gaps between positives and negatives. Under these conditions, retrieval success depends more on a model's general representational capacity than on balanced cross-modal reasoning. As a result, models may still perform well even when over-relying on a single modality, as also demonstrated in Figure~\ref{fig:problem}. This explains why focus imbalances are less visible in standard benchmarks but become critical in the \dataname~benchmark with more difficult settings.

\textbf{Question \#2: Are there hard cases that models with balanced focuses may still fail to handle?}

In existing CIR literature, the difficulty of a sample is typically determined by how challenging it is for a model to distinguish the positive from its corresponding negatives. Accordingly, semantic discrepancies among candidates play a central role.

Negatives with small semantic discrepancies from the positive can be categorized into three types:
\begin{itemize}
\item Negatives that are consistent with the query image but inconsistent with the query text. Models that overfocus on visual information are prone to fail in such cases, and focus balancing directly mitigates this issue.
\item Negatives that are consistent with the query text but inconsistent with the query image. Similarly, models that overfocus on textual information tend to fail, and focus balancing is again beneficial.
\item Negatives that are semantically consistent with both the query image and the query text, but with lower overall consistency than the positive. Even models with balanced focuses may struggle in these cases.
\end{itemize}

Therefore, balanced focus does not guarantee success on all hard cases, particularly those involving ``balanced but relatively weaker semantic consistency''. However, the proposed \dataname~workflow explicitly addresses such cases through two mechanisms:
\begin{itemize}
\item \textbf{Image-augmented negative generation.} As stated in Section~\ref{sec:data} in the main paper, a specific type of negatives are generated using instructions that integrate both image and text semantics. As typical examples with ``balanced but relatively weaker consistency'', such negatives help models learn finer-grained distinctions. Note that since the integrated instructions naturally align more closely with the textual modality, such negatives are especially effective in mitigating overfocuses on text.
\item \textbf{Filtering of original positives.} As stated in Section~\ref{sec:data} in the main paper, original positives that are not highly consistent with the query are reclassified as negatives. Such negatives are another type of examples with ``balanced but relatively weaker consistency'', and further improve model robustness in the above hard scenarios.
\end{itemize}

\textbf{Question \#3: Are there counter-examples where focus balancing negatively impacts performance?}

As stated in Section~\ref{sec:intro} in the main paper, models with imbalanced focuses may still produce correct retrieval results in simple settings, primarily due to large semantic gaps among candidates. Importantly, encouraging balanced focuses in such cases does not degrade performance, as balanced reasoning fundamentally outperforms simpler retrieval behavior.

Another specific scenario arises when test data contain high-amplitude noise that affects one modality. In such cases, models with balanced focuses may be more susceptible to misleading signals. In this work, we attribute such failures to data quality issues rather than an inherent drawback of focus balancing. Under realistic and well-constructed CIR settings, balanced focuses remain desirable properties for robust composed retrieval.

\section{Quality Analyses of the \dataname~Benchmark and Finetuning Dataset}\label{app:data}

The high quality of the \dataname~benchmark and finetuning dataset can be reflected from the following perspectives:
\begin{itemize}
\item \textbf{Query quality:} The proposed workflow preserves the original queries from existing CIR triplets. Since these triplets are sampled from established CIR datasets, the resulting queries naturally satisfy the requirements of composed image retrieval. In this work, queries are collected from multiple widely used open-source CIR datasets, providing a reliable foundation for query quality.
\item \textbf{Candidate quality:} High query quality alone does not guarantee high candidate quality, particularly due to the query-target consistency issue stated in Section~\ref{sec:data} in the main paper. For datasets constructed via similarity-driven image pairing, we synthesize a more consistent positive image for each triplet, yielding substantially improved alignment between queries and targets compared to the original data. In addition, the image editing and generation models employed in the workflow produce visually coherent images that closely follow the instruction prompts, ensuring both semantic consistency and visual fidelity among candidates.
\item \textbf{Query coverage (diversity):} Similarly, the query coverage of \dataname~is inherited from the source CIR triplets. By aggregating and randomly sampling triplets from three large-scale datasets, the resulting data exhibit broad coverage over query types and semantic variations. In addition, query coverage naturally increases with data scale, and can be adjusted to suit specific experimental or application settings.
\item \textbf{Candidate coverage (diversity):} As presented in Table~\ref{tab:benchmark} in the main paper, the proposed benchmark achieves a substantially higher candidate-to-query ratio than existing benchmarks. This indicates low candidate reuse and semantic redundancy, leading to greater diversity among candidates. Such diversity is particularly important for evaluating and diagnosing focus imbalances under challenging retrieval scenarios.
\end{itemize}
In summary, beyond increasing task difficulty and exposing focus imbalances, the proposed \dataname~workflow also improves the overall quality of existing CIR triplets by enhancing query-target consistency and candidate diversity. These properties make the resulting benchmark and finetuning dataset suitable for both diagnostic evaluation and robust model training.

\begin{figure*}[!t]
\centering
\includegraphics[width=0.9\linewidth]{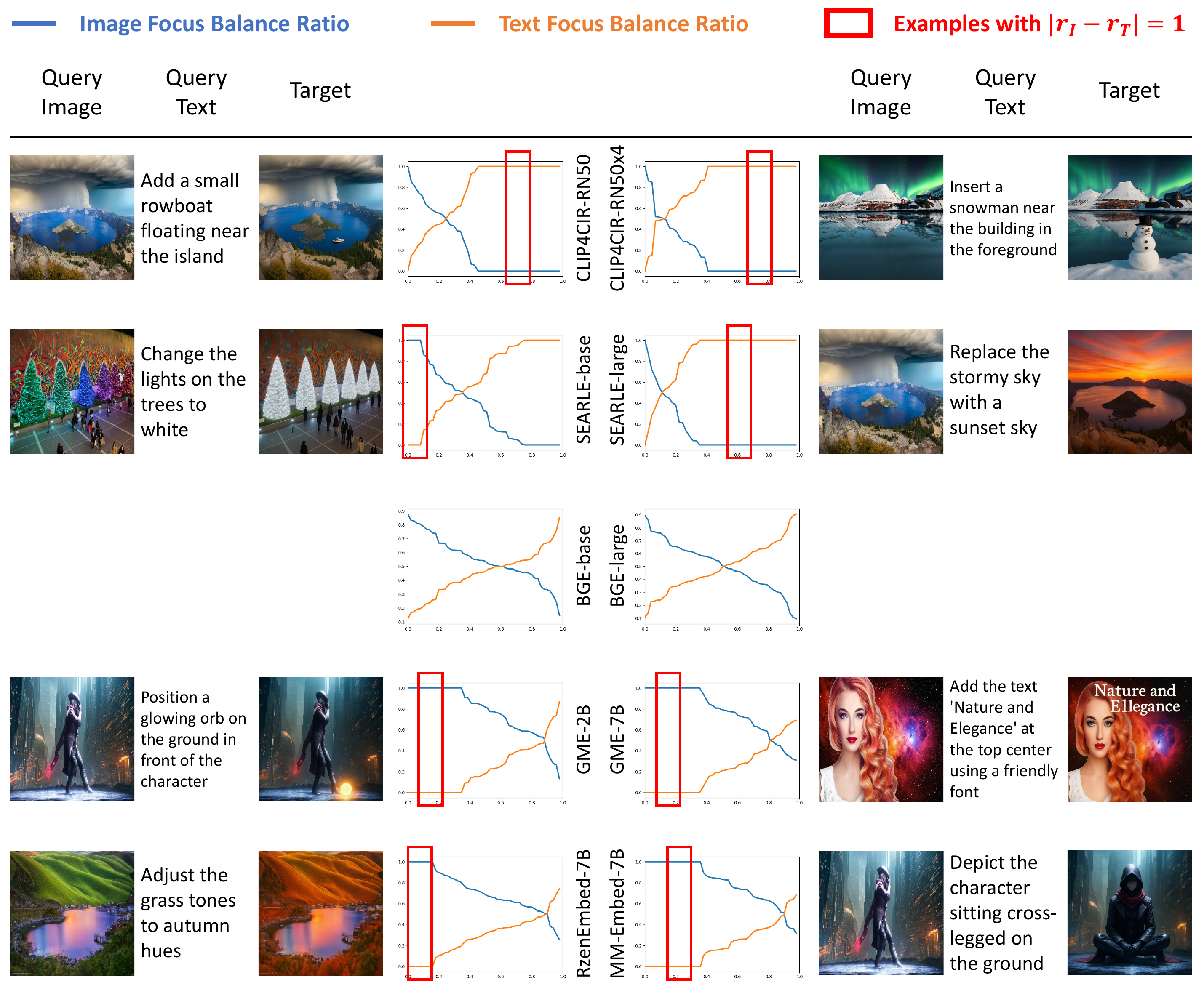}
\caption{Visualizations of image and text focus balance ratios, and examples with semantics of a specific modality completely ignored by the models.}
\vspace{-0.2cm}
\label{fig:focus}
\end{figure*}

\section{Focus Visualizations and Qualitative Analyses of CIR Models on the \dataname~Benchmark}\label{app:focus_vis}

In this section, we randomly select 50 samples from the \dataname~benchmark, compute sample-wise focus balance ratios, and perform qualitative analyses of focus-related model behavior.

Figure~\ref{fig:focus} shows the behavior of the evaluated models on the samples. For better clarity in visualizations, image focus balance ratios $r_I$ are sorted in a descending order.

Results show that focus imbalances are prevalent across all evaluated models, as reflected by the gaps between the image (blue) and text (orange) focus ratio curves. In addition, most models exhibit focus imbalances 1 on specific samples, as highlighted by red boxes, indicating that the models make retrieval decisions with one input modality completely ignored. Note that the BGE models~\cite{zhou2025megapairs} do not exhibit complete focus imbalances on the selected samples. Nevertheless, the gaps between their image and text focus ratios still suggest non-negligible imbalance.

For samples with focus imbalances 1, we present one representative visualization per model in Figure~\ref{fig:focus}. For example, for the CLIP4CIR-RN50~\cite{baldrati2023composed} model, the image semantics of the samples highlighted in the red box are completely ignored. In such cases, model decisions are primarily based on textual semantics (e.g., ``ignoring the image content, and searching for images with \textit{a small rowboat floating near the island}''). While correct retrieval may still occur when the dataset distribution is favorable, such behavior indicates the presence of shortcut strategies learned during training.

Similar patterns can be observed across other models, suggesting that reliance on image-only or text-only shortcuts is a common outcome when training data and task settings are relatively simple. In contrast, the proposed \dataname~workflow mitigates such shortcut behavior by hard negative construction. As illustrated in Figure~\ref{fig:problem} in the main paper, models trained with the augmented data are exposed to hard cases where neither modality alone is sufficient, thereby discouraging shortcut learning and encouraging more balanced cross-modal reasoning.

\input{tables/params}

\section{Detailed Configurations of Model Finetuning}\label{app:exp_details}

As stated in Section~\ref{sec:finetune} in the main paper, we apply an SFT-style finetuning procedure to all evaluated models. The lightweight finetuning strategy has the following properties:
\begin{itemize}
\item The finetuning dataset is relatively small in scale (5K-50K samples in this paper), and consists exclusively of the proposed augmented data.
\item Most model parameters are frozen during finetuning, with only a small proportion ($<$2\% in most cases in this paper) of modules set trainable.
\item The models are trained for a small number of iterations (1 epoch in this paper).
\end{itemize}

For CLIP-based models, except for the BGE model family (which are implemented using the \texttt{AutoModel} wrapper from the \texttt{transformers} library), only the last few layers are set trainable. All VLM-based models are finetuned using LoRA modules with different ranks. Details of the trainable parameters for each model are listed in Table~\ref{tab:params}.

During finetuning, for each training sample, we randomly select three in-sample hard negatives from the corresponding local candidate pool. We adopt a standard contrastive loss with temperature set to 0.07, applied to both in-batch (common-case) negatives and in-sample (hard-case) negatives with different weights. The weight for in-batch negatives is fixed to 1, while the weight for in-sample negatives is gradually increased from 0.2 to 2 over the first 15\% of finetuning steps. This schedule encourages the model to progressively focus on harder cases without destabilizing early training. In addition, we further incorporate a distillation loss to mitigate catastrophic forgetting. For each model, the pretrained checkpoint serves as the teacher, and the distillation loss regularizes the discrepancies between the logits produced by the teacher and the finetuned student:
\begin{equation}
\mathcal{L}_d=\tau^2\cdot\text{KL}(\text{softmax}(\frac{p_\text{student}}{\tau}),\text{softmax}(\frac{p_\text{teacher}}{\tau}))
\end{equation}
where the temperature $\tau$ is set to 2. The weight of the distillation loss is set to $10^3$ in all experiments.

All models are finetuned using 8x NVIDIA H20 (96GB) GPUs. We adopt a global batch size of 64 = per-GPU batch size (4) $\times$ number of GPUs (8) $\times$ number of gradient accumulation steps (2). A cosine learning rate scheduler is applied with an initial learning rate of $10^{-4}$.

\section{Comprehensive Finetuned Model Performance Analyses}\label{app:finetune_all}

\input{tables/finetune_recall_all}
\input{tables/finetune_focus_all}
\input{tables/finetune_recall_zeroshot_all}

\subsection{Finetuned Performance of the VLM-Based Models}\label{app:finetune_vlm_all}

The finetuned benchmark performance, focus imbalances, and zero-shot hard-case performance gains of the evaluated VLM-based models under LoRA ranks 2, 4, 8, and 16 are respectively reported in Table~\ref{tab:finetune_recall_all},~\ref{tab:finetune_focus_all}, and~\ref{tab:finetune_recall_zeroshot_all}, which are respectively the extensions of Table~\ref{tab:finetune_recall},~\ref{tab:finetune_focus}, and~\ref{tab:finetune_recall_zeroshot} in the main paper.

Across all LoRA configurations, the observed trends are consistent with those reported in the main experiments. Finetuned models exhibit improved hard-case performance and reduced focus imbalances, while maintaining competitive performance on standard benchmarks. These results further validate the effectiveness of the proposed \dataname~workflow in encouraging balanced cross-modal reasoning and improving robustness under challenging retrieval scenarios. Moreover, results indicate that the consistent model performance improvements across different LoRA ranks are not sensitive to specific parameterization choices.

\subsection{Finetuned Performance of the CLIP-Based Models}\label{app:finetune_clip}

\input{tables/finetune_recall_clip}

We apply the identical finetuning procedure to CLIP-based CIR models, and report the results in Table~\ref{tab:finetune_recall_clip}. Compared to VLM-based models, most CLIP-based models show less pronounced performance gains, and in some cases exhibit performance drops on standard benchmarks. We discuss several possible contributing factors below:
\begin{itemize}
\item \textbf{Differences in benchmark objectives.}
As stated in Section~\ref{sec:data} in the main paper, the \dataname~benchmark is designed to remove potential ``shortcuts'' that existing CIR models may exploit during training. In contrast, many standard benchmarks implicitly reward such shortcuts (e.g., background or low-level correlation cues).
When finetuned to abandon shortcut strategies, models may experience apparent performance drops on easier benchmarks, even though their robustness under challenging cases improves. Notably, as shown in the main experiments, this trade-off disappears for higher-capacity VLM-based models, demonstrating the effectiveness of the \dataname~workflow.
\item \textbf{Relatively limited model capacity.} Due to the highly-entangled multi-modal feature spaces of CLIP~\cite{radford2021learning} models~\cite{chen2024hibug}, CLIP-based CIR models may struggle to simultaneously capture fine-grained semantic distinctions required by the hard negatives in \dataname~and preserve the coarse-grained features effective for standard benchmarks. In contrast, higher-capacity VLM-based models are better equipped to handle this trade-off.
\item \textbf{Constraints of the finetuning strategy.} Unlike VLM-based models, CLIP-based models are generally not finetuned with LoRA. In addition, for CLIP4CIR~\cite{baldrati2023composed} and SEARLE~\cite{baldrati2023zero}, the selection of trainable layers is heuristic rather than optimized for maximal performance gains. In this work, we focus on evaluating the quality and effectiveness of the proposed data augmentation rather than exploring model-specific finetuning strategies. The consistent improvements of VLM-based models demonstrate the effectiveness of the data itself. For CLIP-based models, the proposed data can be combined with more advanced or tailored finetuning techniques to improve their specific capabilities according to task requirements.
\end{itemize}

\subsection{Ablation Studies Targeting Focus Imbalances}\label{app:focus_all}

\input{tables/finetune_focus_neg_ratio_all}

In this subsection, we conduct ablation studies similar to those in Section~\ref{sec:ablation} in the main paper, with a specific focus on the changes in focus imbalances. Table~\ref{tab:finetune_focus_neg_ratio_all} reports the focus imbalance values under varying data scales and negative ratios.

Across most configurations, finetuning with the proposed hard negatives consistently reduces focus imbalances compared to the baseline data without augmentation. These results provide further evidence that the crafted hard negatives in \dataname~are effective in encouraging more balanced cross-modal focuses.

\section{Benchmark-Wise Analyses of Models' Zero-Shot Performance Gains}\label{app:zeroshot_analysis}

\input{tables/finetune_recall_common_gme}
\input{tables/finetune_recall_common_rzen_mmembed}

Table~\ref{tab:finetune_recall} in the main paper and Table~\ref{tab:finetune_recall_all} report the performance gains of the models finetuned with the proposed data on standard CIR benchmarks. In this section, we provide more detailed benchmark-wise analyses of their zero-shot hard-case performance, complementing the results reported in Table~\ref{tab:finetune_recall_zeroshot} in the main paper and Table~\ref{tab:finetune_recall_zeroshot_all}.

The standard benchmarks adopted in this work exhibit distinct characteristics:
\begin{itemize}
\item CIRR~\cite{liu2021image}: A real-world dataset that additionally defines a candidate subset for each query, where the candidates are semantically closer to one another than those in the global pool. Therefore, subset retrieval on CIRR (measured by Rs@1) poses a more challenging setting than standard global retrieval tasks, making it particularly suitable for evaluating hard-case robustness.
\item FashionIQ~\cite{wu2021fashion}: A fashion clothes dataset with relatively narrow visual categories and limited diversity. Consequently, it exhibits noticeable distribution differences from other benchmarks and from the datasets used to construct \dataname.
\item GeneCIS~\cite{vaze2023genecis}: A real-world dataset that separates queries into object-related and attribute-related categories.
\end{itemize}
Note that CIRCO~\cite{baldrati2023zero} is another widely-adopted CIR benchmark. However, its multi-positive retrieval formulation is not directly compatible with our augmentation strategy, which constructs hard negatives that are semantically close to a single positive and the query. For this reason, we do not include CIRCO in our evaluation.

The respective zero-shot performance gains of the models on the standard benchmarks are reported in Table~\ref{tab:finetune_recall_common_gme} and~\ref{tab:finetune_recall_common_rzen_mmembed}. Results are highly consistent with the benchmark characteristics outlined above:
\begin{itemize}
\item \textbf{Stronger gains on CIRR Rs@1.} Performance improvements on CIRR Rs@1 are generally larger than those observed on other benchmark metrics. Since \dataname~is explicitly designed to target hard cases, models show the most pronounced gains on the most challenging evaluation settings. This provides additional evidence for the effectiveness of the proposed workflow.
\item \textbf{Higher gains on CIRR than on other benchmarks.} Performance improvements on CIRR are generally larger than those on other benchmarks.
CIRR focuses on general image retrieval and shares greater similarity with the base datasets used in constructing \dataname~(MegaPairs~\cite{zhou2025megapairs}, SynthTriplets18M~\cite{gu2023compodiff}, and GPT-Image-Edit-1.5M~\cite{wang2025gpt}). In contrast, models' zero-shot performance on FashionIQ and GeneCIS shows more variability, particularly for GME models~\cite{zhang2024gme}. This behavior likely reflects the increased specialization of these benchmarks and their distribution differences from \dataname, rather than a limitation of the proposed augmentation strategy.
\end{itemize}

\section{Computation Analysis}\label{app:computation}

The \methodname~focus interpretation method stated in Section~\ref{sec:interpretation} involves iterative pruning over multi-modal tokens, resulting in a combinatorial growth in time complexity. In this section, we provide a quantitative analysis of the computational costs of the process and discuss its practical implications.

As shown in Figure~\ref{fig:refinement}, the majority of computations occur in the model inferences during the state validation stage. In the $k$th iteration of the refinement process, each input state has $k-1$ tokens pruned. Following the beam-search strategy stated in Section~\ref{sec:refinement}, we set the maximum number of valid states in each iteration to a constant $w$. Consequently, at most $w\times(n_I+n_T-k+1)$ new raw states can be derived in the $k$th iteration.

Therefore, the maximum number of model inferences required by the entire refinement process can therefore be upper-bounded as:
\begin{equation}
\mathcal{N}=\sum_{k=1}^{n_I+n_T}w\times(n_I+n_T-k+1)=\frac{w}{2}\times(n_I+n_T+1)\times(n_I+n_T)
\end{equation}
which grows quadratically with respect to the total number of image and text tokens. This bound corresponds to a worst-case scenario in which all generated states remain valid and the refinement proceeds until all tokens are pruned. In practice, the refinement process typically terminates earlier, resulting in fewer iterations and inferences.

In our implementation, we adopt the SAM-ViT-Huge~\cite{kirillov2023segment} image segmentation model with moderate granularity (\texttt{points\_per\_side=15}) to determine image tokens. On average, the constructed dataset contains 15.7 image tokens and 10.3 text tokens per sample. With $w=5$, the theoretical worst case corresponds to approximately 1690 model inferences per sample.

\input{tables/computation}

Table~\ref{tab:computation} reports the average inference time measured on a batch size of 12 using 1x NVIDIA H20 GPU with \texttt{bfloat16} precision. Based on the above analyses, the estimated worst-case runtime ranges from 7.0 to 181.7 seconds per sample, depending on the parameter scales of the models. While such computational costs may be non-negligible in extreme cases, \methodname~is explicitly designed for offline model diagnosis and guidance for the data augmentation workflow instead of real-time analyses over large-scale datasets. As a result, the computational overhead of \methodname~does not affect the efficiency of model training, evaluation, or deployment, and remains acceptable within its intended diagnostic scope.

%% file: tables/params.tex
\begin{table}[!t]
\centering
\small
\caption{Details of the trainable parameters of the tested CIR models.}
\begin{tabular}{cccc}
\toprule
Model & LoRA Ranks & Trainable Param Count (Proportion) & Trainable Param Range \\
\midrule
\multicolumn{4}{c}{CLIP-based Models} \\
\midrule
\rule[-1.5ex]{0pt}{4.5ex} CLIP4CIR-RN50 & N/A & 75.5M (29.8\%) & \multirow{2}{*}{\makecell[l]{\texttt{combiner.dynamic\_scalar} \\ \texttt{combiner.output\_layer} \\ \texttt{clip\_model.logit\_scale}}} \\
\cmidrule{1-3}
\rule[-1.5ex]{0pt}{4.5ex} CLIP4CIR-RN50x4 & N/A & 29.5M (12.4\%) & \\
\midrule
\rule{0pt}{2ex} SEARLE-base & N/A & 1.0M (0.7\%) & \multirow{2}{*}{\makecell[l]{\texttt{searle.layers.6} \\ \texttt{clip\_model.logit\_scale}}} \\
\cmidrule{1-3}
\rule{0pt}{2ex} SEARLE-large & N/A & 2.4M (0.5\%) & \\
\midrule
\multirow{4}{*}{BGE-base} & 2 & 0.2M (0.2\%) & \multirow{8}{*}{\makecell[l]{\texttt{q\_proj} \\ \texttt{k\_proj} \\ \texttt{v\_proj} \\ \texttt{out\_proj}}} \\
& 4 & 0.5M (0.3\%) & \\
& 8 & 1.0M (0.7\%) & \\
& 16 & 2.0M (1.3\%) & \\
\cmidrule{1-3}
\multirow{4}{*}{BGE-large} & 2 & 0.5M (0.1\%) & \\
& 4 & 1.1M (0.3\%) & \\
& 8 & 2.2M (0.5\%) & \\
& 16 & 4.3M (1.0\%) & \\
\midrule
\multicolumn{4}{c}{VLM-based Models} \\
\midrule
\multirow{4}{*}{GME-2B} & 2 & 2.3M (0.1\%) & \multirow{16}{*}{\makecell[l]{\texttt{q\_proj} \\ \texttt{k\_proj} \\ \texttt{v\_proj} \\ \texttt{o\_proj} \\ \texttt{gate\_proj} \\ \texttt{up\_proj} \\ \texttt{down\_proj}}} \\
& 4 & 4.6M (0.2\%) & \\
& 8 & 9.2M (0.4\%) & \\
& 16 & 18.5M (0.8\%) & \\
\cmidrule{1-3}
\multirow{4}{*}{GME-7B} & 2 & 5.0M (0.1\%) & \\
& 4 & 10.1M (0.2\%) & \\
& 8 & 20.2M (0.3\%) & \\
& 16 & 40.4M (0.5\%) & \\
\cmidrule{1-3}
\multirow{4}{*}{RzenEmbed-7B} & 2 & 5.0M (0.1\%) & \\
& 4 & 10.1M (0.2\%) & \\
& 8 & 20.2M (0.3\%) & \\
& 16 & 40.4M (0.5\%) & \\
\cmidrule{1-3}
\multirow{4}{*}{MM-Embed-7B} & 2 & 5.5M (0.1\%) & \\
& 4 & 11.1M (0.2\%) & \\
& 8 & 22.2M (0.3\%) & \\
& 16 & 44.3M (0.5\%) & \\
\bottomrule
\end{tabular}
\vspace{-0.2cm}
\label{tab:params}
\end{table}

%% file: tables/finetune_recall_all.tex
\begin{table}[!t]
\centering
\small
\caption{Finetuned performance of the VLM-based CIR models trained with different data scales and LoRA ranks on the standard and the \dataname~benchmarks (extension of Table~\ref{tab:finetune_recall} in the main paper). Higher is better.}
\begin{tabular}{c|ccc|ccc}
\toprule
\multicolumn{7}{c}{Metrics: CIRR R@1 / CIRR Rs@1 / FashionIQ R@10 / GeneCIS Rs@1 $\ ||\ $ Average (Gain) $\ ||\ $ \dataname~Rs@1 (Gain)} \\
\midrule
Data / Rank & \multicolumn{3}{c|}{GME-2B} & \multicolumn{3}{c}{GME-7B} \\
\midrule
Pretrained & 32.8 / 41.5 / 39.8 / 21.1 & 33.8 & 29.7 & 35.2 / 44.3 / 43.3 / 21.7 & 36.1 & 32.9 \\
\midrule
5K / 2 & 36.8 / 48.3 / 40.0 / 20.9 & 36.5 (\textcolor{green}{+2.7}) & 37.6 (\textcolor{green}{+7.9}) & 41.1 / 51.1 / 43.2 / 21.7 & 39.3 (\textcolor{green}{+3.1}) & 42.2 (\textcolor{green}{+9.3}) \\
10K / 2 & 37.1 / 48.4 / 39.7 / 21.3 & 36.6 (\textcolor{green}{+2.8}) & 37.7 (\textcolor{green}{+8.0}) & 44.1 / 55.3 / 43.2 / 21.8 & 41.1 (\textcolor{green}{+5.0}) & 42.7 (\textcolor{green}{+9.8}) \\
20K / 2 & 35.3 / 45.6 / 39.9 / 20.5 & 35.3 (\textcolor{green}{+1.5}) & 36.6 (\textcolor{green}{+6.9}) & 43.5 / 54.6 / 43.2 / 21.7 & 40.7 (\textcolor{green}{+4.6}) & 44.2 (\textcolor{green}{+11.3}) \\
50K / 2 & 36.8 / 47.7 / 40.1 / 20.8 & 36.4 (\textcolor{green}{+2.6}) & 37.9 (\textcolor{green}{+8.2}) & 44.5 / 56.2 / 43.0 / 21.4 & 41.3 (\textcolor{green}{+5.1}) & 46.6 (\textcolor{green}{+13.7}) \\
\midrule
5K / 4 & 35.5 / 45.7 / 40.1 / 21.3 & 35.6 (\textcolor{green}{+1.9}) & 35.8 (\textcolor{green}{+6.1}) & 43.6 / 54.8 / 43.4 / 22.0 & 40.9 (\textcolor{green}{+4.8}) & 43.1 (\textcolor{green}{+10.2}) \\
10K / 4 & 35.7 / 45.7 / 39.8 / 20.4 & 35.4 (\textcolor{green}{+1.6}) & 36.1 (\textcolor{green}{+6.4}) & 42.9 / 53.8 / 43.2 / 21.5 & 40.3 (\textcolor{green}{+4.2}) & 43.4 (\textcolor{green}{+10.5}) \\
20K / 4 & 37.1 / 48.2 / 40.1 / 21.1 & 36.6 (\textcolor{green}{+2.8}) & 40.7 (\textcolor{green}{+11.0}) & 42.5 / 52.9 / 43.2 / 21.9 & 40.1 (\textcolor{green}{+4.0}) & 45.0 (\textcolor{green}{+12.1}) \\
50K / 4 & 36.5 / 47.8 / 39.6 / 20.5 & 36.1 (\textcolor{green}{+2.3}) & 38.8 (\textcolor{green}{+9.1}) & 46.4 / 59.2 / 43.3 / 21.9 & 42.7 (\textcolor{green}{+6.6}) & 47.3 (\textcolor{green}{+14.4}) \\
\midrule
5K / 8 & 35.8 / 46.8 / 40.0 / 21.0 & 35.9 (\textcolor{green}{+2.1}) & 37.5 (\textcolor{green}{+7.8}) & 43.1 / 54.1 / 43.1 / 22.1 & 40.6 (\textcolor{green}{+4.5}) & 43.8 (\textcolor{green}{+10.9}) \\
10K / 8 & 37.1 / 48.5 / 39.9 / 20.8 & 36.6 (\textcolor{green}{+2.8}) & 38.4 (\textcolor{green}{+8.7}) & 44.0 / 55.6 / 43.1 / 21.7 & 41.1 (\textcolor{green}{+5.0}) & 45.6 (\textcolor{green}{+12.7}) \\
20K / 8 & 37.1 / 48.7 / 40.0 / 20.7 & 36.6 (\textcolor{green}{+2.8}) & 41.1 (\textcolor{green}{+11.4}) & 44.1 / 55.6 / 43.0 / 21.4 & 41.0 (\textcolor{green}{+4.9}) & 44.9 (\textcolor{green}{+12.0}) \\
50K / 8 & 36.8 / 47.8 / 39.8 / 21.3 & 36.4 (\textcolor{green}{+2.7}) & 41.5 (\textcolor{green}{+11.8}) & 42.8 / 54.1 / 42.5 / 21.4 & 40.2 (\textcolor{green}{+4.1}) & 45.9 (\textcolor{green}{+13.0}) \\
\midrule
5K / 16 & 35.8 / 46.5 / 39.7 / 20.9 & 35.7 (\textcolor{green}{+1.9}) & 36.9 (\textcolor{green}{+7.2}) & 40.7 / 50.3 / 42.9 / 21.3 & 38.8 (\textcolor{green}{+2.7}) & 43.4 (\textcolor{green}{+10.5}) \\
10K / 16 & 35.1 / 45.8 / 39.5 / 21.0 & 35.3 (\textcolor{green}{+1.6}) & 36.3 (\textcolor{green}{+6.6}) & 40.2 / 48.7 / 42.9 / 21.4 & 38.3 (\textcolor{green}{+2.2}) & 42.3 (\textcolor{green}{+9.4}) \\
20K / 16 & 36.2 / 46.5 / 39.6 / 20.3 & 35.7 (\textcolor{green}{+1.9}) & 39.7 (\textcolor{green}{+10.0}) & 42.3 / 52.9 / 43.4 / 21.6 & 40.0 (\textcolor{green}{+3.9}) & 45.4 (\textcolor{green}{+12.5}) \\
50K / 16 & 37.5 / 48.6 / 39.8 / 20.9 & 36.7 (\textcolor{green}{+2.9}) & 43.5 (\textcolor{green}{+13.8}) & 42.0 / 51.9 / 42.8 / 21.3 & 39.5 (\textcolor{green}{+3.4}) & 46.1 (\textcolor{green}{+13.2}) \\
\midrule
Data / Rank & \multicolumn{3}{c|}{RzenEmbed-7B} & \multicolumn{3}{c}{MM-Embed-7B} \\
\midrule
Pretrained & 40.1 / 47.5 / 41.2 / 21.4 & 37.6 & 39.3 & 24.5 / 27.2 / 38.2 / 18.9 & 27.2 & 26.7 \\
\midrule
5K / 2 & 41.5 / 49.8 / 41.3 / 22.1 & 38.6 (\textcolor{green}{+1.1}) & 42.2 (\textcolor{green}{+2.9}) & 26.6 / 30.0 / 38.3 / 19.1 & 28.5 (\textcolor{green}{+1.3}) & 30.9 (\textcolor{green}{+4.2}) \\
10K / 2 & 43.7 / 53.5 / 40.7 / 23.6 & 40.4 (\textcolor{green}{+2.8}) & 44.7 (\textcolor{green}{+5.4}) & 28.3 / 32.2 / 38.6 / 19.1 & 29.6 (\textcolor{green}{+2.4}) & 34.2 (\textcolor{green}{+7.5}) \\
20K / 2 & 45.5 / 55.9 / 40.8 / 23.0 & 41.3 (\textcolor{green}{+3.7}) & 44.1 (\textcolor{green}{+4.8}) & 30.2 / 35.0 / 39.0 / 20.7 & 31.2 (\textcolor{green}{+4.0}) & 35.0 (\textcolor{green}{+8.3}) \\
50K / 2 & 44.6 / 54.8 / 40.6 / 22.9 & 40.7 (\textcolor{green}{+3.2}) & 45.4 (\textcolor{green}{+6.1}) & 31.3 / 36.8 / 38.9 / 20.6 & 31.9 (\textcolor{green}{+4.7}) & 38.5 (\textcolor{green}{+11.8}) \\
\midrule
5K / 4 & 42.2 / 51.1 / 41.5 / 22.1 & 39.2 (\textcolor{green}{+1.7}) & 42.4 (\textcolor{green}{+3.1}) & 24.5 / 27.4 / 38.8 / 18.2 & 27.2 (+0.0) & 30.2 (\textcolor{green}{+3.5}) \\
10K / 4 & 41.6 / 49.8 / 41.6 / 22.3 & 38.9 (\textcolor{green}{+1.3}) & 42.5 (\textcolor{green}{+3.2}) & 27.4 / 30.8 / 38.4 / 20.6 & 29.3 (\textcolor{green}{+2.1}) & 32.8 (\textcolor{green}{+6.1}) \\
20K / 4 & 40.5 / 47.4 / 41.4 / 21.5 & 37.7 (\textcolor{green}{+0.1}) & 42.8 (\textcolor{green}{+3.5}) & 31.3 / 36.6 / 38.4 / 21.4 & 31.9 (\textcolor{green}{+4.7}) & 35.7 (\textcolor{green}{+9.0}) \\
50K / 4 & 43.0 / 50.8 / 40.9 / 22.4 & 39.3 (\textcolor{green}{+1.7}) & 45.5 (\textcolor{green}{+6.2}) & 26.3 / 29.2 / 37.8 / 20.1 & 28.4 (\textcolor{green}{+1.2}) & 35.7 (\textcolor{green}{+9.0}) \\
\midrule
5K / 8 & 42.2 / 51.0 / 41.7 / 22.6 & 39.4 (\textcolor{green}{+1.8}) & 42.5 (\textcolor{green}{+3.2}) & 30.3 / 34.7 / 38.8 / 20.9 & 31.2 (\textcolor{green}{+4.0}) & 33.0 (\textcolor{green}{+6.3}) \\
10K / 8 & 42.4 / 51.3 / 41.5 / 23.0 & 39.5 (\textcolor{green}{+1.9}) & 42.5 (\textcolor{green}{+3.2}) & 27.9 / 31.9 / 38.6 / 20.5 & 29.7 (\textcolor{green}{+2.5}) & 34.0 (\textcolor{green}{+7.3}) \\
20K / 8 & 39.5 / 46.6 / 41.5 / 22.3 & 37.5 (\textcolor{red}{-0.1}) & 42.5 (\textcolor{green}{+3.2}) & 26.8 / 30.2 / 38.2 / 20.6 & 28.9 (\textcolor{green}{+1.7}) & 36.9 (\textcolor{green}{+10.2}) \\
50K / 8 & 41.0 / 48.5 / 41.3 / 21.4 & 38.0 (\textcolor{green}{+0.5}) & 44.6 (\textcolor{green}{+5.3}) & 29.6 / 34.5 / 38.2 / 20.6 & 30.7 (\textcolor{green}{+3.5}) & 38.4 (\textcolor{green}{+11.7}) \\
\midrule
5K / 16 & 40.4 / 48.4 / 41.1 / 22.0 & 38.0 (\textcolor{green}{+0.4}) & 42.9 (\textcolor{green}{+3.6}) & 29.5 / 33.9 / 38.7 / 19.9 & 30.5 (\textcolor{green}{+3.3}) & 34.1 (\textcolor{green}{+7.4}) \\
10K / 16 & 40.2 / 47.8 / 42.0 / 22.9 & 38.2 (\textcolor{green}{+0.6}) & 44.0 (\textcolor{green}{+4.7}) & 33.9 / 40.0 / 39.5 / 21.8 & 33.8 (\textcolor{green}{+6.6}) & 38.0 (\textcolor{green}{+11.3}) \\
20K / 16 & 42.4 / 51.0 / 41.3 / 22.4 & 39.3 (\textcolor{green}{+1.7}) & 44.5 (\textcolor{green}{+5.2}) & 27.0 / 30.3 / 37.0 / 19.8 & 28.5 (\textcolor{green}{+1.3}) & 34.6 (\textcolor{green}{+7.9}) \\
50K / 16 & 43.1 / 51.8 / 41.6 / 23.2 & 39.9 (\textcolor{green}{+2.4}) & 46.5 (\textcolor{green}{+7.2}) & 24.6 / 27.1 / 38.2 / 18.4 & 27.1 (\textcolor{red}{-0.1}) & 35.7 (\textcolor{green}{+9.0}) \\
\bottomrule
\end{tabular}
\vspace{-0.2cm}
\label{tab:finetune_recall_all}
\end{table}

%% file: tables/finetune_focus_all.tex
\begin{table}[t]
\centering
\small
\caption{Finetuned focus imbalances of the VLM-based CIR models trained with different data scales and LoRA ranks (extension of Table~\ref{tab:finetune_focus} in the main paper). Lower is better.}
\begin{tabular}{c|cccc|cccc}
\toprule
\multicolumn{9}{c}{Metrics: Focus Imbalance (Reduction)} \\
\midrule
Data & 5K & 10K & 20K & 50K & 5K & 10K & 20K & 50K \\
\midrule
Rank & \multicolumn{4}{c|}{GME-2B (Pretrained: 0.54)} & \multicolumn{4}{c}{GME-7B (Pretrained: 0.55)} \\
\midrule
2 & 0.51 (\textcolor{green}{-0.03}) & 0.58 (\textcolor{red}{+0.04}) & 0.21 (\textcolor{green}{-0.33}) & 0.28 (\textcolor{green}{-0.26}) & 0.54 (\textcolor{green}{-0.01}) & 0.54 (\textcolor{green}{-0.02}) & 0.23 (\textcolor{green}{-0.33}) & 0.13 (\textcolor{green}{-0.42}) \\
4 & 0.56 (\textcolor{red}{+0.02}) & 0.40 (\textcolor{green}{-0.14}) & 0.17 (\textcolor{green}{-0.38}) & 0.32 (\textcolor{green}{-0.22}) & 0.54 (\textcolor{green}{-0.01}) & 0.54 (\textcolor{green}{-0.02}) & 0.23 (\textcolor{green}{-0.33}) & 0.13 (\textcolor{green}{-0.42}) \\
8 & 0.58 (\textcolor{red}{+0.04}) & 0.16 (\textcolor{green}{-0.38}) & 0.20 (\textcolor{green}{-0.34}) & 0.17 (\textcolor{green}{-0.37}) & 0.53 (\textcolor{green}{-0.03}) & 0.26 (\textcolor{green}{-0.29}) & 0.27 (\textcolor{green}{-0.29}) & 0.31 (\textcolor{green}{-0.24}) \\
16 & 0.56 (\textcolor{red}{+0.02}) & 0.34 (\textcolor{green}{-0.20}) & 0.32 (\textcolor{green}{-0.22}) & 0.27 (\textcolor{green}{-0.27}) & 0.56 (-0.00) & 0.22 (\textcolor{green}{-0.33}) & 0.35 (\textcolor{green}{-0.21}) & 0.28 (\textcolor{green}{-0.28}) \\
\midrule
Rank & \multicolumn{4}{c|}{RzenEmbed-7B (Pretrained: 0.43)} & \multicolumn{4}{c}{MM-Embed-7B (Pretrained: 0.60)} \\
\midrule
2 & 0.44 (\textcolor{red}{+0.01}) & 0.23 (\textcolor{green}{-0.21}) & 0.10 (\textcolor{green}{-0.33}) & 0.11 (\textcolor{green}{-0.32}) & 0.56 (\textcolor{green}{-0.04}) & 0.05 (\textcolor{green}{-0.55}) & 0.06 (\textcolor{green}{-0.54}) & 0.16 (\textcolor{green}{-0.44}) \\
4 & 0.44 (\textcolor{red}{+0.01}) & 0.08 (\textcolor{green}{-0.35}) & 0.21 (\textcolor{green}{-0.22}) & 0.04 (\textcolor{green}{-0.39}) & 0.67 (\textcolor{red}{+0.07}) & 0.17 (\textcolor{green}{-0.43}) & 0.05 (\textcolor{green}{-0.55}) & 0.14 (\textcolor{green}{-0.45}) \\
8 & 0.46 (\textcolor{red}{+0.03}) & 0.19 (\textcolor{green}{-0.24}) & 0.12 (\textcolor{green}{-0.31}) & 0.04 (\textcolor{green}{-0.39}) & 0.62 (\textcolor{red}{+0.02}) & 0.01 (\textcolor{green}{-0.59}) & 0.10 (\textcolor{green}{-0.50}) & 0.06 (\textcolor{green}{-0.54}) \\
16 & 0.42 (\textcolor{green}{-0.01}) & 0.18 (\textcolor{green}{-0.25}) & 0.10 (\textcolor{green}{-0.33}) & 0.06 (\textcolor{green}{-0.37}) & 0.07 (\textcolor{green}{-0.53}) & 0.02 (\textcolor{green}{-0.57}) & 0.17 (\textcolor{green}{-0.43}) & 0.19 (\textcolor{green}{-0.41}) \\
\bottomrule
\end{tabular}
\vspace{-0.2cm}
\label{tab:finetune_focus_all}
\end{table}

%% file: tables/finetune_recall_zeroshot_all.tex
\begin{table}[t]
\centering
\small
\caption{Finetuned zero-shot performance of the VLM-based CIR models trained with different data scales and LoRA ranks on the \methodname-augmented CIRR benchmark (extension of Table~\ref{tab:finetune_recall_zeroshot} in the main paper). Higher is better.}
\begin{tabular}{c|cccc|cccc}
\toprule
\multicolumn{9}{c}{Metrics: \methodname-CIRR Rs@1 (Gain)} \\
\midrule
Data & 5K & 10K & 20K & 50K & 5K & 10K & 20K & 50K \\
\midrule
Rank & \multicolumn{4}{c|}{GME-2B (Pretrained: 28.5)} & \multicolumn{4}{c}{GME-7B (Pretrained: 25.0)} \\
\midrule
2 & 29.8 (\textcolor{green}{+1.3}) & 29.6 (\textcolor{green}{+1.1}) & 28.0 (\textcolor{red}{-0.5}) & 30.2 (\textcolor{green}{+1.7}) & 27.4 (\textcolor{green}{+2.4}) & 27.7 (\textcolor{green}{+2.7}) & 27.3 (\textcolor{green}{+2.3}) & 29.0 (\textcolor{green}{+4.0}) \\
4 & 29.2 (\textcolor{green}{+0.7}) & 28.7 (\textcolor{green}{+0.2}) & 29.7 (\textcolor{green}{+1.2}) & 30.0 (\textcolor{green}{+1.5}) & 27.3 (\textcolor{green}{+2.3}) & 28.7 (\textcolor{green}{+3.7}) & 29.4 (\textcolor{green}{+4.4}) & 30.1 (\textcolor{green}{+5.1}) \\
8 & 28.8 (\textcolor{green}{+0.3}) & 29.4 (\textcolor{green}{+0.9}) & 30.3 (\textcolor{green}{+1.8}) & 30.5 (\textcolor{green}{+2.0}) & 29.0 (\textcolor{green}{+4.0}) & 27.7 (\textcolor{green}{+2.7}) & 28.7 (\textcolor{green}{+3.7}) & 28.5 (\textcolor{green}{+3.5}) \\
16 & 29.3 (\textcolor{green}{+0.8}) & 28.0 (\textcolor{red}{-0.5}) & 29.3 (\textcolor{green}{+0.8}) & 31.5 (\textcolor{green}{+3.0}) & 27.7 (\textcolor{green}{+2.7}) & 28.5 (\textcolor{green}{+3.5}) & 28.9 (\textcolor{green}{+3.9}) & 30.4 (\textcolor{green}{+5.4}) \\
\midrule
Rank & \multicolumn{4}{c|}{RzenEmbed-7B (Pretrained: 30.1)} & \multicolumn{4}{c}{MM-Embed-7B (Pretrained: 27.5)} \\
\midrule
2 & 30.5 (\textcolor{green}{+0.4}) & 31.5 (\textcolor{green}{+1.4}) & 30.4 (\textcolor{green}{+0.3}) & 33.2 (\textcolor{green}{+3.1}) & 29.4 (\textcolor{green}{+1.9}) & 31.5 (\textcolor{green}{+4.0}) & 29.5 (\textcolor{green}{+2.0}) & 33.3 (\textcolor{green}{+5.8}) \\
4 & 30.8 (\textcolor{green}{+0.7}) & 32.0 (\textcolor{green}{+1.9}) & 32.2 (\textcolor{green}{+2.1}) & 34.0 (\textcolor{green}{+3.9}) & 30.7 (\textcolor{green}{+3.2}) & 30.8 (\textcolor{green}{+3.3}) & 31.9 (\textcolor{green}{+4.4}) & 30.9 (\textcolor{green}{+3.4}) \\
8 & 30.5 (\textcolor{green}{+0.4}) & 30.8 (\textcolor{green}{+0.7}) & 33.0 (\textcolor{green}{+2.9}) & 33.8 (\textcolor{green}{+3.7}) & 31.4 (\textcolor{green}{+3.9}) & 30.9 (\textcolor{green}{+3.4}) & 32.4 (\textcolor{green}{+4.9}) & 31.8 (\textcolor{green}{+4.3}) \\
16 & 32.3 (\textcolor{green}{+2.2}) & 32.5 (\textcolor{green}{+2.4}) & 32.4 (\textcolor{green}{+2.3}) & 32.0 (\textcolor{green}{+1.9}) & 30.3 (\textcolor{green}{+2.8}) & 32.6 (\textcolor{green}{+5.1}) & 33.0 (\textcolor{green}{+5.5}) & 33.3 (\textcolor{green}{+5.8}) \\
\bottomrule
\end{tabular}
\vspace{-0.2cm}
\label{tab:finetune_recall_zeroshot_all}
\end{table}

%% file: tables/finetune_recall_clip.tex
\begin{table}[t]
\centering
\small
\caption{Finetuned performance of the CLIP-based CIR models trained with different data scales and LoRA ranks on the standard and the \dataname~benchmarks. Higher is better.}
\begin{tabular}{c|cccc|cccc}
\toprule
\multicolumn{9}{c}{Metrics: Average (Gain) / \dataname~Rs@1 (Gain)} \\
\midrule
Data & 5K & 10K & 20K & 50K & 5K & 10K & 20K & 50K \\
\midrule
Rank & \multicolumn{4}{c|}{CLIP4CIR-RN50 (Pretrained: 24.9 / 33.4)} & \multicolumn{4}{c}{CLIP4CIR-RN50x4 (Pretrained: 34.2 / 41.4)} \\
\midrule
\multirow{2}{*}{N/A} & 20.3 (\textcolor{red}{-4.6}) & 20.2 (\textcolor{red}{-4.6}) & 17.6 (\textcolor{red}{-7.3}) & 19.1 (\textcolor{red}{-5.7}) & 27.8 (\textcolor{red}{-6.4}) & 26.5 (\textcolor{red}{-7.6}) & 25.6 (\textcolor{red}{-8.6}) & 25.4 (\textcolor{red}{-8.8}) \\
& 42.2 (\textcolor{green}{+8.8}) & 47.0 (\textcolor{green}{+13.6}) & 41.7 (\textcolor{green}{+8.3}) & 48.4 (\textcolor{green}{+15.0}) & 50.7 (\textcolor{green}{+9.3}) & 51.0 (\textcolor{green}{+9.6}) & 53.8 (\textcolor{green}{+12.4}) & 51.4 (\textcolor{green}{+10.0}) \\
\midrule
Rank & \multicolumn{4}{c|}{SEARLE-base (Pretrained: 23.6 / 29.3)} & \multicolumn{4}{c}{SEARLE-large (Pretrained: 24.7 / 30.2)} \\
\midrule
\multirow{2}{*}{N/A} & 23.0 (\textcolor{red}{-0.6}) & 22.7 (\textcolor{red}{-0.9}) & 22.9 (\textcolor{red}{-0.7}) & 22.5 (\textcolor{red}{-1.0}) & 24.7 (+0.0) & 24.4 (\textcolor{red}{-0.3}) & 23.9 (\textcolor{red}{-0.7}) & 23.3 (\textcolor{red}{-1.4}) \\
& 31.5 (\textcolor{green}{+2.2}) & 32.2 (\textcolor{green}{+2.9}) & 35.4 (\textcolor{green}{+6.1}) & 40.0 (\textcolor{green}{+10.7}) & 37.9 (\textcolor{green}{+7.7}) & 39.3 (\textcolor{green}{+9.1}) & 44.9 (\textcolor{green}{+14.7}) & 51.3 (\textcolor{green}{+21.1}) \\
\midrule
Rank & \multicolumn{4}{c|}{BGE-base (Pretrained: 36.5 / 41.6)} & \multicolumn{4}{c}{BGE-large (Pretrained: 38.3 / 38.8)} \\
\midrule
\multirow{2}{*}{2} & 36.1 (\textcolor{red}{-0.3}) & 35.9 (\textcolor{red}{-0.5}) & 35.8 (\textcolor{red}{-0.7}) & 35.0 (\textcolor{red}{-1.4}) & 37.5 (\textcolor{red}{-0.8}) & 37.4 (\textcolor{red}{-0.8}) & 37.6 (\textcolor{red}{-0.7}) & 37.4 (\textcolor{red}{-0.9}) \\
& 42.4 (\textcolor{green}{+0.8}) & 41.8 (\textcolor{green}{+0.2}) & 43.0 (\textcolor{green}{+1.4}) & 44.6 (\textcolor{green}{+3.0}) & 40.7 (\textcolor{green}{+1.9}) & 40.7 (\textcolor{green}{+1.9}) & 40.8 (\textcolor{green}{+2.0}) & 42.3 (\textcolor{green}{+3.5}) \\
\midrule
\multirow{2}{*}{4} & 36.1 (\textcolor{red}{-0.4}) & 35.6 (\textcolor{red}{-0.9}) & 35.9 (\textcolor{red}{-0.5}) & 35.8 (\textcolor{red}{-0.7}) & 37.6 (\textcolor{red}{-0.6}) & 37.6 (\textcolor{red}{-0.7}) & 37.6 (\textcolor{red}{-0.6}) & 37.3 (\textcolor{red}{-1.0}) \\
& 42.4 (\textcolor{green}{+0.8}) & 42.9 (\textcolor{green}{+1.3}) & 42.3 (\textcolor{green}{+0.7}) & 44.2 (\textcolor{green}{+2.6}) & 41.0 (\textcolor{green}{+2.2}) & 40.8 (\textcolor{green}{+2.0}) & 40.9 (\textcolor{green}{+2.1}) & 42.1 (\textcolor{green}{+3.3}) \\
\midrule
\multirow{2}{*}{8} & 35.9 (\textcolor{red}{-0.6}) & 35.9 (\textcolor{red}{-0.6}) & 35.9 (\textcolor{red}{-0.5}) & 36.0 (\textcolor{red}{-0.5}) & 37.8 (\textcolor{red}{-0.4}) & 38.0 (\textcolor{red}{-0.3}) & 37.6 (\textcolor{red}{-0.7}) & 37.6 (\textcolor{red}{-0.7}) \\
& 41.9 (\textcolor{green}{+0.3}) & 43.6 (\textcolor{green}{+2.0}) & 43.6 (\textcolor{green}{+2.0}) & 44.4 (\textcolor{green}{+2.8}) & 40.2 (\textcolor{green}{+1.4}) & 41.0 (\textcolor{green}{+2.2}) & 41.9 (\textcolor{green}{+3.1}) & 42.2 (\textcolor{green}{+3.4}) \\
\midrule
\multirow{2}{*}{16} & 36.0 (\textcolor{red}{-0.4}) & 35.9 (\textcolor{red}{-0.6}) & 35.7 (\textcolor{red}{-0.7}) & 35.8 (\textcolor{red}{-0.6}) & 37.9 (\textcolor{red}{-0.3}) & 37.6 (\textcolor{red}{-0.7}) & 37.6 (\textcolor{red}{-0.7}) & 37.3 (\textcolor{red}{-1.0}) \\
& 42.4 (\textcolor{green}{+0.8}) & 43.5 (\textcolor{green}{+1.9}) & 42.8 (\textcolor{green}{+1.2}) & 45.3 (\textcolor{green}{+3.7}) & 40.6 (\textcolor{green}{+1.8}) & 40.9 (\textcolor{green}{+2.1}) & 41.5 (\textcolor{green}{+2.7}) & 41.4 (\textcolor{green}{+2.6}) \\
\bottomrule
\end{tabular}
\vspace{-0.2cm}
\label{tab:finetune_recall_clip}
\end{table}

%% file: tables/finetune_focus_neg_ratio_all.tex
\begin{table}[!t]
\centering
\small
\caption{Finetuned (with 2-rank LoRA) focus imbalances of the VLM-based CIR models trained with different data scales and negative ratios. Lower is better.}
\begin{tabular}{c|cccc|cccc}
\toprule
\multicolumn{9}{c}{Metrics: Focus Imbalance (Reduction)} \\
\midrule
Neg & 5K & 10K & 20K & 50K & 5K & 10K & 20K & 50K \\
\cmidrule{2-9}
Ratio & \multicolumn{4}{c|}{GME-2B (Pretrained: 0.54)} & \multicolumn{4}{c}{GME-7B (Pretrained: 0.55)} \\
\midrule
0\% & 0.58 & 0.53 & 0.15 & 0.28 & 0.49 & 0.53 & 0.32 & 0.28 \\
\midrule
25\% & 0.52 (\textcolor{green}{-0.05}) & 0.37 (\textcolor{green}{-0.16}) & 0.28 (\textcolor{red}{+0.13}) & 0.20 (\textcolor{green}{-0.08}) & 0.52 (\textcolor{red}{+0.03}) & 0.55 (\textcolor{red}{+0.02}) & 0.28 (\textcolor{green}{-0.04}) & 0.12 (\textcolor{green}{-0.17}) \\
50\% & 0.51 (\textcolor{green}{-0.07}) & 0.58 (\textcolor{red}{+0.05}) & 0.21 (\textcolor{red}{+0.06}) & 0.28 (\textcolor{red}{+0.01}) & 0.54 (\textcolor{red}{+0.05}) & 0.54 (\textcolor{red}{+0.01}) & 0.23 (\textcolor{green}{-0.09}) & 0.13 (\textcolor{green}{-0.15}) \\
75\% & 0.48 (\textcolor{green}{-0.10}) & 0.30 (\textcolor{green}{-0.22}) & 0.23 (\textcolor{red}{+0.08}) & 0.25 (\textcolor{green}{-0.03}) & 0.46 (\textcolor{green}{-0.03}) & 0.36 (\textcolor{green}{-0.17}) & 0.29 (\textcolor{green}{-0.02}) & 0.26 (\textcolor{green}{-0.03}) \\
100\% & 0.45 (\textcolor{green}{-0.13}) & 0.53 (-0.00) & 0.17 (\textcolor{red}{+0.02}) & 0.21 (\textcolor{green}{-0.07}) & 0.53 (\textcolor{red}{+0.04}) & 0.19 (\textcolor{green}{-0.34}) & 0.12 (\textcolor{green}{-0.20}) & 0.32 (\textcolor{red}{+0.03}) \\
\midrule
& \multicolumn{4}{c|}{RzenEmbed-7B (Pretrained: 0.43)} & \multicolumn{4}{c}{MM-Embed-7B (Pretrained: 0.60)} \\
\midrule
0\% & 0.45 & 0.46 & 0.34 & 0.22 & 0.68 & 0.18 & 0.17 & 0.24 \\
\midrule
25\% & 0.45 (-0.00) & 0.54 (\textcolor{red}{+0.08}) & 0.05 (\textcolor{green}{-0.29}) & 0.22 (-0.00) & 0.02 (\textcolor{green}{-0.66}) & 0.12 (\textcolor{green}{-0.07}) & 0.12 (\textcolor{green}{-0.05}) & 0.07 (\textcolor{green}{-0.16}) \\
50\% & 0.44 (\textcolor{green}{-0.01}) & 0.23 (\textcolor{green}{-0.24}) & 0.10 (\textcolor{green}{-0.24}) & 0.11 (\textcolor{green}{-0.10}) & 0.56 (\textcolor{green}{-0.12}) & 0.05 (\textcolor{green}{-0.14}) & 0.06 (\textcolor{green}{-0.11}) & 0.16 (\textcolor{green}{-0.08}) \\
75\% & 0.44 (\textcolor{green}{-0.01}) & 0.04 (\textcolor{green}{-0.42}) & 0.05 (\textcolor{green}{-0.29}) & 0.13 (\textcolor{green}{-0.09}) & 0.57 (\textcolor{green}{-0.11}) & 0.01 (\textcolor{green}{-0.18}) & 0.04 (\textcolor{green}{-0.13}) & 0.04 (\textcolor{green}{-0.19}) \\
100\% & 0.44 (\textcolor{green}{-0.01}) & 0.07 (\textcolor{green}{-0.40}) & 0.24 (\textcolor{green}{-0.11}) & 0.01 (\textcolor{green}{-0.20}) & 0.63 (\textcolor{green}{-0.05}) & 0.08 (\textcolor{green}{-0.11}) & 0.04 (\textcolor{green}{-0.14}) & 0.11 (\textcolor{green}{-0.12}) \\
\bottomrule
\end{tabular}
\vspace{-0.2cm}
\label{tab:finetune_focus_neg_ratio_all}
\end{table}

%% file: tables/finetune_recall_common_gme.tex
\begin{table}[h]
\centering
\small
\caption{Finetuned performance of the VLM-based GME models trained with different data scales and LoRA ranks on the standard benchmarks (extension of Table~\ref{tab:finetune_recall} in the main paper). Higher is better.}
\begin{tabular}{ccccc}
\toprule
Data / Rank & CIRR R@1 & CIRR Rs@1 & FashionIQ R@1 & GeneCIS Rs@1 \\
\midrule
\multicolumn{5}{c}{GME-2B} \\
\midrule
Pretrained & 32.8 & 41.5 & 39.8 & 21.1 \\
\midrule
5K / 2 & 36.8 (\textcolor{green}{+4.0}) & 48.3 (\textcolor{green}{+6.8}) & 40.0 (\textcolor{green}{+0.2}) & 20.9 (\textcolor{red}{-0.2}) \\
10K / 2 & 37.1 (\textcolor{green}{+4.4}) & 48.4 (\textcolor{green}{+6.9}) & 39.7 (\textcolor{red}{-0.1}) & 21.3 (\textcolor{green}{+0.2}) \\
20K / 2 & 35.3 (\textcolor{green}{+2.5}) & 45.6 (\textcolor{green}{+4.1}) & 39.9 (+0.0) & 20.5 (\textcolor{red}{-0.6}) \\
50K / 2 & 36.8 (\textcolor{green}{+4.1}) & 47.7 (\textcolor{green}{+6.3}) & 40.1 (\textcolor{green}{+0.3}) & 20.8 (\textcolor{red}{-0.3}) \\
5K / 4 & 35.5 (\textcolor{green}{+2.8}) & 45.7 (\textcolor{green}{+4.2}) & 40.1 (\textcolor{green}{+0.3}) & 21.3 (\textcolor{green}{+0.2}) \\
10K / 4 & 35.7 (\textcolor{green}{+2.9}) & 45.7 (\textcolor{green}{+4.3}) & 39.8 (+0.0) & 20.4 (\textcolor{red}{-0.8}) \\
20K / 4 & 37.1 (\textcolor{green}{+4.4}) & 48.2 (\textcolor{green}{+6.7}) & 40.1 (\textcolor{green}{+0.3}) & 21.1 (+0.0) \\
50K / 4 & 36.5 (\textcolor{green}{+3.8}) & 47.8 (\textcolor{green}{+6.3}) & 39.6 (\textcolor{red}{-0.2}) & 20.5 (\textcolor{red}{-0.6}) \\
5K / 8 & 35.8 (\textcolor{green}{+3.0}) & 46.8 (\textcolor{green}{+5.4}) & 40.0 (\textcolor{green}{+0.2}) & 21.0 (\textcolor{red}{-0.1}) \\
10K / 8 & 37.1 (\textcolor{green}{+4.3}) & 48.5 (\textcolor{green}{+7.1}) & 39.9 (\textcolor{green}{+0.1}) & 20.8 (\textcolor{red}{-0.3}) \\
20K / 8 & 37.1 (\textcolor{green}{+4.3}) & 48.7 (\textcolor{green}{+7.2}) & 40.0 (\textcolor{green}{+0.2}) & 20.7 (\textcolor{red}{-0.4}) \\
50K / 8 & 36.8 (\textcolor{green}{+4.1}) & 47.8 (\textcolor{green}{+6.4}) & 39.8 (+0.0) & 21.3 (\textcolor{green}{+0.2}) \\
5K / 16 & 35.8 (\textcolor{green}{+3.0}) & 46.5 (\textcolor{green}{+5.0}) & 39.7 (\textcolor{red}{-0.1}) & 20.9 (\textcolor{red}{-0.2}) \\
10K / 16 & 35.1 (\textcolor{green}{+2.4}) & 45.8 (\textcolor{green}{+4.3}) & 39.5 (\textcolor{red}{-0.3}) & 21.0 (\textcolor{red}{-0.2}) \\
20K / 16 & 36.2 (\textcolor{green}{+3.5}) & 46.5 (\textcolor{green}{+5.0}) & 39.6 (\textcolor{red}{-0.2}) & 20.3 (\textcolor{red}{-0.8}) \\
50K / 16 & 37.5 (\textcolor{green}{+4.7}) & 48.6 (\textcolor{green}{+7.2}) & 39.8 (+0.0) & 20.9 (\textcolor{red}{-0.3}) \\
\midrule
Average Gain & \textcolor{green}{+3.6} & \textcolor{green}{+5.8} & +0.0 & \textcolor{red}{-0.3} \\
\midrule
\multicolumn{5}{c}{GME-7B} \\
\midrule
Pretrained & 35.2 & 44.3 & 43.3 & 21.7 \\
\midrule
5K / 2 & 41.1 (\textcolor{green}{+5.9}) & 51.1 (\textcolor{green}{+6.7}) & 43.2 (\textcolor{red}{-0.1}) & 21.7 (+0.0) \\
10K / 2 & 44.1 (\textcolor{green}{+8.9}) & 55.3 (\textcolor{green}{+11.0}) & 43.2 (\textcolor{red}{-0.1}) & 21.8 (\textcolor{green}{+0.2}) \\
20K / 2 & 43.5 (\textcolor{green}{+8.3}) & 54.6 (\textcolor{green}{+10.2}) & 43.2 (\textcolor{red}{-0.1}) & 21.7 (+0.0) \\
50K / 2 & 44.5 (\textcolor{green}{+9.3}) & 56.2 (\textcolor{green}{+11.8}) & 43.0 (\textcolor{red}{-0.3}) & 21.4 (\textcolor{red}{-0.3}) \\
5K / 4 & 43.6 (\textcolor{green}{+8.4}) & 54.8 (\textcolor{green}{+10.5}) & 43.4 (+0.0) & 22.0 (\textcolor{green}{+0.3}) \\
10K / 4 & 42.9 (\textcolor{green}{+7.7}) & 53.8 (\textcolor{green}{+9.5}) & 43.2 (\textcolor{red}{-0.2}) & 21.5 (\textcolor{red}{-0.2}) \\
20K / 4 & 42.5 (\textcolor{green}{+7.3}) & 52.9 (\textcolor{green}{+8.6}) & 43.2 (\textcolor{red}{-0.1}) & 21.9 (\textcolor{green}{+0.3}) \\
50K / 4 & 46.4 (\textcolor{green}{+11.2}) & 59.2 (\textcolor{green}{+14.9}) & 43.3 (+0.0) & 21.9 (\textcolor{green}{+0.2}) \\
5K / 8 & 43.1 (\textcolor{green}{+7.9}) & 54.1 (\textcolor{green}{+9.7}) & 43.1 (\textcolor{red}{-0.2}) & 22.1 (\textcolor{green}{+0.4}) \\
10K / 8 & 44.0 (\textcolor{green}{+8.9}) & 55.6 (\textcolor{green}{+11.2}) & 43.1 (\textcolor{red}{-0.2}) & 21.7 (+0.0) \\
20K / 8 & 44.1 (\textcolor{green}{+8.9}) & 55.6 (\textcolor{green}{+11.2}) & 43.0 (\textcolor{red}{-0.3}) & 21.4 (\textcolor{red}{-0.3}) \\
50K / 8 & 42.8 (\textcolor{green}{+7.7}) & 54.1 (\textcolor{green}{+9.8}) & 42.5 (\textcolor{red}{-0.8}) & 21.4 (\textcolor{red}{-0.3}) \\
5K / 16 & 40.7 (\textcolor{green}{+5.6}) & 50.3 (\textcolor{green}{+5.9}) & 42.9 (\textcolor{red}{-0.4}) & 21.3 (\textcolor{red}{-0.4}) \\
10K / 16 & 40.2 (\textcolor{green}{+5.0}) & 48.7 (\textcolor{green}{+4.4}) & 42.9 (\textcolor{red}{-0.5}) & 21.4 (\textcolor{red}{-0.3}) \\
20K / 16 & 42.3 (\textcolor{green}{+7.2}) & 52.9 (\textcolor{green}{+8.6}) & 43.4 (+0.0) & 21.6 (\textcolor{red}{-0.1}) \\
50K / 16 & 42.0 (\textcolor{green}{+6.8}) & 51.9 (\textcolor{green}{+7.5}) & 42.8 (\textcolor{red}{-0.5}) & 21.3 (\textcolor{red}{-0.4}) \\
\midrule
Average Gain & \textcolor{green}{+7.8} & \textcolor{green}{+9.5} & \textcolor{red}{-0.2} & \textcolor{red}{-0.1} \\
\bottomrule
\end{tabular}
\vspace{-0.2cm}
\label{tab:finetune_recall_common_gme}
\end{table}

%% file: tables/finetune_recall_common_rzen_mmembed.tex
\begin{table}[h]
\centering
\small
\caption{Finetuned performance of the VLM-based RzenEmbed-7B and MM-Embed-7B models trained with different data scales and LoRA ranks on the standard benchmarks (extension of Table~\ref{tab:finetune_recall} in the main paper). Higher is better.}
\begin{tabular}{ccccc}
\toprule
Data / Rank & CIRR R@1 & CIRR Rs@1 & FashionIQ R@1 & GeneCIS Rs@1 \\
\midrule
\multicolumn{5}{c}{RzenEmbed-7B} \\
\midrule
Pretrained & 40.1 & 47.5 & 41.2 & 21.4 \\
\midrule
5K / 2 & 41.5 (\textcolor{green}{+1.3}) & 49.8 (\textcolor{green}{+2.3}) & 41.3 (\textcolor{green}{+0.1}) & 22.1 (\textcolor{green}{+0.7}) \\
10K / 2 & 43.7 (\textcolor{green}{+3.5}) & 53.5 (\textcolor{green}{+6.0}) & 40.7 (\textcolor{red}{-0.5}) & 23.6 (\textcolor{green}{+2.2}) \\
20K / 2 & 45.5 (\textcolor{green}{+5.3}) & 55.9 (\textcolor{green}{+8.4}) & 40.8 (\textcolor{red}{-0.4}) & 23.0 (\textcolor{green}{+1.5}) \\
50K / 2 & 44.6 (\textcolor{green}{+4.5}) & 54.8 (\textcolor{green}{+7.3}) & 40.6 (\textcolor{red}{-0.6}) & 22.9 (\textcolor{green}{+1.4}) \\
5K / 4 & 42.2 (\textcolor{green}{+2.1}) & 51.1 (\textcolor{green}{+3.6}) & 41.5 (\textcolor{green}{+0.2}) & 22.1 (\textcolor{green}{+0.7}) \\
10K / 4 & 41.6 (\textcolor{green}{+1.5}) & 49.8 (\textcolor{green}{+2.3}) & 41.6 (\textcolor{green}{+0.4}) & 22.3 (\textcolor{green}{+0.9}) \\
20K / 4 & 40.5 (\textcolor{green}{+0.4}) & 47.4 (\textcolor{red}{-0.1}) & 41.4 (\textcolor{green}{+0.2}) & 21.5 (\textcolor{green}{+0.1}) \\
50K / 4 & 43.0 (\textcolor{green}{+2.8}) & 50.8 (\textcolor{green}{+3.3}) & 40.9 (\textcolor{red}{-0.3}) & 22.4 (\textcolor{green}{+1.0}) \\
5K / 8 & 42.2 (\textcolor{green}{+2.1}) & 51.0 (\textcolor{green}{+3.5}) & 41.7 (\textcolor{green}{+0.4}) & 22.6 (\textcolor{green}{+1.2}) \\
10K / 8 & 42.4 (\textcolor{green}{+2.2}) & 51.3 (\textcolor{green}{+3.8}) & 41.5 (\textcolor{green}{+0.2}) & 23.0 (\textcolor{green}{+1.5}) \\
20K / 8 & 39.5 (\textcolor{red}{-0.6}) & 46.6 (\textcolor{red}{-0.9}) & 41.5 (\textcolor{green}{+0.3}) & 22.3 (\textcolor{green}{+0.9}) \\
50K / 8 & 41.0 (\textcolor{green}{+0.8}) & 48.5 (\textcolor{green}{+1.1}) & 41.3 (\textcolor{green}{+0.1}) & 21.4 (+0.0) \\
5K / 16 & 40.4 (\textcolor{green}{+0.3}) & 48.4 (\textcolor{green}{+0.9}) & 41.1 (\textcolor{red}{-0.1}) & 22.0 (\textcolor{green}{+0.6}) \\
10K / 16 & 40.2 (\textcolor{green}{+0.1}) & 47.8 (\textcolor{green}{+0.3}) & 42.0 (\textcolor{green}{+0.7}) & 22.9 (\textcolor{green}{+1.4}) \\
20K / 16 & 42.4 (\textcolor{green}{+2.2}) & 51.0 (\textcolor{green}{+3.5}) & 41.3 (\textcolor{green}{+0.1}) & 22.4 (\textcolor{green}{+1.0}) \\
50K / 16 & 43.1 (\textcolor{green}{+2.9}) & 51.8 (\textcolor{green}{+4.3}) & 41.6 (\textcolor{green}{+0.4}) & 23.2 (\textcolor{green}{+1.7}) \\
\midrule
Average Gain & \textcolor{green}{+2.0} & \textcolor{green}{+3.1} & \textcolor{green}{+0.1} & \textcolor{green}{+1.1} \\
\midrule
\multicolumn{5}{c}{MM-Embed-7B} \\
\midrule
Pretrained & 24.5 & 27.2 & 38.2 & 18.9 \\
\midrule
5K / 2 & 26.6 (\textcolor{green}{+2.0}) & 30.0 (\textcolor{green}{+2.8}) & 38.3 (\textcolor{green}{+0.1}) & 19.1 (\textcolor{green}{+0.3}) \\
10K / 2 & 28.3 (\textcolor{green}{+3.7}) & 32.2 (\textcolor{green}{+5.0}) & 38.6 (\textcolor{green}{+0.4}) & 19.1 (\textcolor{green}{+0.3}) \\
20K / 2 & 30.2 (\textcolor{green}{+5.6}) & 35.0 (\textcolor{green}{+7.8}) & 39.0 (\textcolor{green}{+0.8}) & 20.7 (\textcolor{green}{+1.8}) \\
50K / 2 & 31.3 (\textcolor{green}{+6.7}) & 36.8 (\textcolor{green}{+9.5}) & 38.9 (\textcolor{green}{+0.7}) & 20.6 (\textcolor{green}{+1.8}) \\
5K / 4 & 24.5 (+0.0) & 27.4 (\textcolor{green}{+0.2}) & 38.8 (\textcolor{green}{+0.6}) & 18.2 (\textcolor{red}{-0.7}) \\
10K / 4 & 27.4 (\textcolor{green}{+2.8}) & 30.8 (\textcolor{green}{+3.6}) & 38.4 (\textcolor{green}{+0.2}) & 20.6 (\textcolor{green}{+1.7}) \\
20K / 4 & 31.3 (\textcolor{green}{+6.8}) & 36.6 (\textcolor{green}{+9.4}) & 38.4 (\textcolor{green}{+0.2}) & 21.4 (\textcolor{green}{+2.6}) \\
50K / 4 & 26.3 (\textcolor{green}{+1.8}) & 29.2 (\textcolor{green}{+2.0}) & 37.8 (\textcolor{red}{-0.4}) & 20.1 (\textcolor{green}{+1.2}) \\
5K / 8 & 30.3 (\textcolor{green}{+5.8}) & 34.7 (\textcolor{green}{+7.5}) & 38.8 (\textcolor{green}{+0.7}) & 20.9 (\textcolor{green}{+2.0}) \\
10K / 8 & 27.9 (\textcolor{green}{+3.3}) & 31.9 (\textcolor{green}{+4.7}) & 38.6 (\textcolor{green}{+0.5}) & 20.5 (\textcolor{green}{+1.7}) \\
20K / 8 & 26.8 (\textcolor{green}{+2.2}) & 30.2 (\textcolor{green}{+3.0}) & 38.2 (+0.0) & 20.6 (\textcolor{green}{+1.7}) \\
50K / 8 & 29.6 (\textcolor{green}{+5.1}) & 34.5 (\textcolor{green}{+7.3}) & 38.2 (\textcolor{green}{+0.1}) & 20.6 (\textcolor{green}{+1.7}) \\
5K / 16 & 29.5 (\textcolor{green}{+5.0}) & 33.9 (\textcolor{green}{+6.7}) & 38.7 (\textcolor{green}{+0.5}) & 19.9 (\textcolor{green}{+1.0}) \\
10K / 16 & 33.9 (\textcolor{green}{+9.4}) & 40.0 (\textcolor{green}{+12.8}) & 39.5 (\textcolor{green}{+1.3}) & 21.8 (\textcolor{green}{+3.0}) \\
20K / 16 & 27.0 (\textcolor{green}{+2.5}) & 30.3 (\textcolor{green}{+3.1}) & 37.0 (\textcolor{red}{-1.2}) & 19.8 (\textcolor{green}{+1.0}) \\
50K / 16 & 24.6 (\textcolor{green}{+0.1}) & 27.1 (\textcolor{red}{-0.1}) & 38.2 (+0.0) & 18.4 (\textcolor{red}{-0.5}) \\
\midrule
Average Gain & \textcolor{green}{+3.9} & \textcolor{green}{+5.3} & \textcolor{green}{+0.3} & \textcolor{green}{+1.3} \\
\bottomrule
\end{tabular}
\vspace{-0.2cm}
\label{tab:finetune_recall_common_rzen_mmembed}
\end{table}

%% file: tables/computation.tex
\begin{table}[!t]
\centering
\small
\caption{Inference time of the tested CIR models on a batch of size 12 on one single NVIDIA H20 GPU with \texttt{bfloat16} precision.}
\begin{tabular}{cccccc}
\toprule
Model & CLIP4CIR-RN50 & SEARLE-base & BGE-base & GME-2B & RzenEmbed-7B \\
\midrule
Time / s & 0.24 & 0.23 & 0.05 & 0.53 & 0.90 \\
\midrule
Model & CLIP4CIR-RN50x4 & SEARLE-large & BGE-large & GME-7B & MM-Embed-7B \\
\midrule
Time / s & 0.25 & 0.29 & 0.06 & 0.89 & 1.29 \\
\bottomrule
\end{tabular}
\vspace{-0.2cm}
\label{tab:computation}
\end{table}

%% file: main.bib
@article{focalogic,
  title={Focalogic: Logic-based interpretation of visual model decisions}, 
  author={Zhao, Chenchen and Chen, Muxi and Xu, Qiang},
  journal={arXiv preprint arXiv:2601.12049},
  year={2026}
}

@article{hu2021lora,
  title={LoRA: Low-Rank Adaptation of Large Language Models}, 
  author={Edward J. Hu and Yelong Shen and Phillip Wallis and Zeyuan Allen-Zhu and Yuanzhi Li and Shean Wang and Lu Wang and Weizhu Chen},
  journal={arXiv preprint arXiv:2106.09685},
  year={2021}
}

@article{baldrati2023composed,
  title={Composed image retrieval using contrastive learning and task-oriented clip-based features},
  author={Baldrati, Alberto and Bertini, Marco and Uricchio, Tiberio and Del Bimbo, Alberto},
  journal={ACM Transactions on Multimedia Computing, Communications and Applications},
  volume={20},
  number={3},
  pages={1--24},
  year={2023},
  publisher={ACM New York, NY}
}

@article{zhang2024gme,
  title={GME: Improving Universal Multimodal Retrieval by Multimodal LLMs},
  author={Zhang, Xin and Zhang, Yanzhao and Xie, Wen and Li, Mingxin and Dai, Ziqi and Long, Dingkun and Xie, Pengjun and Zhang, Meishan and Li, Wenjie and Zhang, Min},
  journal={arXiv preprint arXiv:2412.16855},
  year={2024}
}

@article{jian2025rzenembed,
  title={Rzenembed: Towards comprehensive multimodal retrieval},
  author={Jian, Weijian and Zhang, Yajun and Liang, Dawei and Xie, Chunyu and He, Yixiao and Leng, Dawei and Yin, Yuhui},
  journal={arXiv preprint arXiv:2510.27350},
  year={2025}
}

@article{lin2024mm,
  title={Mm-embed: Universal multimodal retrieval with multimodal llms},
  author={Lin, Sheng-Chieh and Lee, Chankyu and Shoeybi, Mohammad and Lin, Jimmy and Catanzaro, Bryan and Ping, Wei},
  journal={arXiv preprint arXiv:2411.02571},
  year={2024}
}

@inproceedings{kirillov2023segment,
  title={Segment anything},
  author={Kirillov, Alexander and Mintun, Eric and Ravi, Nikhila and Mao, Hanzi and Rolland, Chloe and Gustafson, Laura and Xiao, Tete and Whitehead, Spencer and Berg, Alexander C and Lo, Wan-Yen and others},
  booktitle={Proceedings of the IEEE/CVF international conference on computer vision},
  pages={4015--4026},
  year={2023}
}

@inproceedings{vo2019composing,
  title={Composing text and image for image retrieval-an empirical odyssey},
  author={Vo, Nam and Jiang, Lu and Sun, Chen and Murphy, Kevin and Li, Li-Jia and Fei-Fei, Li and Hays, James},
  booktitle={Proceedings of the IEEE/CVF conference on computer vision and pattern recognition},
  pages={6439--6448},
  year={2019}
}

@article{wang2025gpt,
  title={Gpt-image-edit-1.5 m: A million-scale, gpt-generated image dataset},
  author={Wang, Yuhan and Yang, Siwei and Zhao, Bingchen and Zhang, Letian and Liu, Qing and Zhou, Yuyin and Xie, Cihang},
  journal={arXiv preprint arXiv:2507.21033},
  year={2025}
}

@inproceedings{zeiler2014visualizing,
  title={Visualizing and understanding convolutional networks},
  author={Zeiler, Matthew D and Fergus, Rob},
  booktitle={European conference on computer vision},
  pages={818--833},
  year={2014},
  organization={Springer}
}

@inproceedings{fong2017interpretable,
  title={Interpretable explanations of black boxes by meaningful perturbation},
  author={Fong, Ruth C and Vedaldi, Andrea},
  booktitle={Proceedings of the IEEE international conference on computer vision},
  pages={3429--3437},
  year={2017}
}

@inproceedings{ribeiro2016should,
  title={"Why should i trust you?" Explaining the predictions of any classifier},
  author={Ribeiro, Marco Tulio and Singh, Sameer and Guestrin, Carlos},
  booktitle={Proceedings of the 22nd ACM SIGKDD international conference on knowledge discovery and data mining},
  pages={1135--1144},
  year={2016}
}

@article{lundberg2017unified,
  title={A unified approach to interpreting model predictions},
  author={Lundberg, Scott M and Lee, Su-In},
  journal={Advances in neural information processing systems},
  volume={30},
  year={2017}
}

@article{simonyan2013deep,
  title={Deep inside convolutional networks: Visualising image classification models and saliency maps},
  author={Simonyan, Karen and Vedaldi, Andrea and Zisserman, Andrew},
  journal={arXiv preprint arXiv:1312.6034},
  year={2013}
}

@inproceedings{selvaraju2017grad,
  title={Grad-cam: Visual explanations from deep networks via gradient-based localization},
  author={Selvaraju, Ramprasaath R and Cogswell, Michael and Das, Abhishek and Vedantam, Ramakrishna and Parikh, Devi and Batra, Dhruv},
  booktitle={Proceedings of the IEEE international conference on computer vision},
  pages={618--626},
  year={2017}
}

@article{chang2018explaining,
  title={Explaining image classifiers by counterfactual generation},
  author={Chang, Chun-Hao and Creager, Elliot and Goldenberg, Anna and Duvenaud, David},
  journal={arXiv preprint arXiv:1807.08024},
  year={2018}
}

@article{bisiani1992beam,
  title={Beam search},
  author={Bisiani, Roberto},
  journal={Encyclopedia of artificial intelligence},
  year={1992},
  publisher={Wiley-Interscience Publication}
}

@article{jiang2024vlm2vec,
  title={Vlm2vec: Training vision-language models for massive multimodal embedding tasks},
  author={Jiang, Ziyan and Meng, Rui and Yang, Xinyi and Yavuz, Semih and Zhou, Yingbo and Chen, Wenhu},
  journal={arXiv preprint arXiv:2410.05160},
  year={2024}
}

@article{gu2023compodiff,
  title={Compodiff: Versatile composed image retrieval with latent diffusion},
  author={Gu, Geonmo and Chun, Sanghyuk and Kim, Wonjae and Jun, HeeJae and Kang, Yoohoon and Yun, Sangdoo},
  journal={arXiv preprint arXiv:2303.11916},
  year={2023}
}

@article{wu2025qwen,
  title={Qwen-image technical report},
  author={Wu, Chenfei and Li, Jiahao and Zhou, Jingren and Lin, Junyang and Gao, Kaiyuan and Yan, Kun and Yin, Sheng-ming and Bai, Shuai and Xu, Xiao and Chen, Yilei and others},
  journal={arXiv preprint arXiv:2508.02324},
  year={2025}
}

@article{yang2025qwen3,
  title={Qwen3 technical report},
  author={Yang, An and Li, Anfeng and Yang, Baosong and Zhang, Beichen and Hui, Binyuan and Zheng, Bo and Yu, Bowen and Gao, Chang and Huang, Chengen and Lv, Chenxu and others},
  journal={arXiv preprint arXiv:2505.09388},
  year={2025}
}

@article{chen2024hibug,
  title={HiBug: on human-interpretable model debug},
  author={Chen, Muxi and Li, Yu and Xu, Qiang},
  journal={Advances in Neural Information Processing Systems},
  volume={36},
  year={2024}
}

@inproceedings{radford2021learning,
  title={Learning transferable visual models from natural language supervision},
  author={Radford, Alec and Kim, Jong Wook and Hallacy, Chris and Ramesh, Aditya and Goh, Gabriel and Agarwal, Sandhini and Sastry, Girish and Askell, Amanda and Mishkin, Pamela and Clark, Jack and others},
  booktitle={International conference on machine learning},
  pages={8748--8763},
  year={2021},
  organization={PmLR}
}

@inproceedings{liu2021image,
  title={Image retrieval on real-life images with pre-trained vision-and-language models},
  author={Liu, Zheyuan and Rodriguez-Opazo, Cristian and Teney, Damien and Gould, Stephen},
  booktitle={Proceedings of the IEEE/CVF international conference on computer vision},
  pages={2125--2134},
  year={2021}
}

@inproceedings{baldrati2022effective,
  title={Effective conditioned and composed image retrieval combining clip-based features},
  author={Baldrati, Alberto and Bertini, Marco and Uricchio, Tiberio and Del Bimbo, Alberto},
  booktitle={Proceedings of the IEEE/CVF conference on computer vision and pattern recognition},
  pages={21466--21474},
  year={2022}
}

@inproceedings{saito2023pic2word,
  title={Pic2word: Mapping pictures to words for zero-shot composed image retrieval},
  author={Saito, Kuniaki and Sohn, Kihyuk and Zhang, Xiang and Li, Chun-Liang and Lee, Chen-Yu and Saenko, Kate and Pfister, Tomas},
  booktitle={Proceedings of the IEEE/CVF Conference on Computer Vision and Pattern Recognition},
  pages={19305--19314},
  year={2023}
}

@inproceedings{vaze2023genecis,
  title={Genecis: A benchmark for general conditional image similarity},
  author={Vaze, Sagar and Carion, Nicolas and Misra, Ishan},
  booktitle={Proceedings of the IEEE/CVF Conference on Computer Vision and Pattern Recognition},
  pages={6862--6872},
  year={2023}
}

@inproceedings{li2022blip,
  title={Blip: Bootstrapping language-image pre-training for unified vision-language understanding and generation},
  author={Li, Junnan and Li, Dongxu and Xiong, Caiming and Hoi, Steven},
  booktitle={International conference on machine learning},
  pages={12888--12900},
  year={2022},
  organization={PMLR}
}

@inproceedings{zhou2025megapairs,
  title={Megapairs: Massive data synthesis for universal multimodal retrieval},
  author={Zhou, Junjie and Xiong, Yongping and Liu, Zheng and Liu, Ze and Xiao, Shitao and Wang, Yueze and Zhao, Bo and Zhang, Chen Jason and Lian, Defu},
  booktitle={Proceedings of the 63rd Annual Meeting of the Association for Computational Linguistics (Volume 1: Long Papers)},
  pages={19076--19095},
  year={2025}
}

@inproceedings{baldrati2023zero,
  title={Zero-shot composed image retrieval with textual inversion},
  author={Baldrati, Alberto and Agnolucci, Lorenzo and Bertini, Marco and Del Bimbo, Alberto},
  booktitle={Proceedings of the IEEE/CVF International Conference on Computer Vision},
  pages={15338--15347},
  year={2023}
}

@article{bai2023qwen,
  title={Qwen technical report},
  author={Bai, Jinze and Bai, Shuai and Chu, Yunfei and Cui, Zeyu and Dang, Kai and Deng, Xiaodong and Fan, Yang and Ge, Wenbin and Han, Yu and Huang, Fei and others},
  journal={arXiv preprint arXiv:2309.16609},
  year={2023}
}

@inproceedings{wu2021fashion,
  title={Fashion iq: A new dataset towards retrieving images by natural language feedback},
  author={Wu, Hui and Gao, Yupeng and Guo, Xiaoxiao and Al-Halah, Ziad and Rennie, Steven and Grauman, Kristen and Feris, Rogerio},
  booktitle={Proceedings of the IEEE/CVF Conference on computer vision and pattern recognition},
  pages={11307--11317},
  year={2021}
}

@inproceedings{isola2015discovering,
  title={Discovering states and transformations in image collections},
  author={Isola, Phillip and Lim, Joseph J and Adelson, Edward H},
  booktitle={Proceedings of the IEEE conference on computer vision and pattern recognition},
  pages={1383--1391},
  year={2015}
}
